\documentclass[11pt]{article}

\usepackage[T1]{fontenc}
\usepackage[utf8]{inputenc}        

\usepackage{amsmath}
\usepackage{amssymb}
\usepackage{newtxtext,newtxmath}
\usepackage[margin=1in]{geometry}
\usepackage{setspace}
\usepackage{microtype}

\usepackage{graphicx}
\usepackage{float}
\usepackage{longtable}
\usepackage{booktabs}              
\usepackage{array}
\usepackage{multirow}
\usepackage{pdflscape}
\usepackage{caption}
\usepackage{amsmath,amssymb}
\usepackage{textcomp}

\usepackage[hidelinks,unicode=true,bookmarks=true]{hyperref}
\usepackage{url}
\usepackage[superscript]{cite}     

\usepackage[affil-it]{authblk}

\usepackage{titlesec}
\titleformat*{\section}{\large\bfseries\sffamily}
\titleformat*{\subsection}{\normalsize\bfseries\sffamily}
\titleformat*{\subsubsection}{\small\bfseries\sffamily}

\newcommand{\D}{\ensuremath{\cdot}}   

\begin{document}

\title{\bfseries Agentic clinical reasoning over longitudinal myeloma records:\\
       a retrospective evaluation against expert consensus}

%
\author[1,2,3]{\mbox{Johannes Moll\thanks{Corresponding author: \texttt{johannes.moll@tum.de}}\textsuperscript{,$\dagger$}}}
\author[2]{\mbox{Jannik L\"ubberstedt\textsuperscript{$\dagger$}}}
\author[4]{\mbox{Christoph Nuernbergk}}
\author[4]{\mbox{Jacob Stroh}}
\author[4]{\mbox{Luisa Mertens}}
\author[4]{\mbox{Anna Purcarea}}
\author[2]{\mbox{Christopher Zirn}}
\author[2]{\mbox{Zeineb Benchaaben}}
\author[1,2]{\mbox{Fabian Drexel}}
\author[2,5]{\mbox{Hartmut H\"antze}}
\author[2,5]{\mbox{Anirudh Narayanan}}
\author[2,5]{\mbox{Friedrich Puttkammer}}
\author[6]{\mbox{Andrei Zhukov}}
\author[7,8]{\mbox{Jacqueline Lammert}}
\author[2]{\mbox{Sebastian Ziegelmayer}}
\author[2]{\mbox{Markus Graf}}
\author[4]{\mbox{Marion H\"ogner}}
\author[2]{\mbox{Marcus Makowski}}
\author[4,9,10,11]{\mbox{Florian Bassermann}}
\author[2]{\mbox{Lisa C. Adams}}
\author[1,12,13]{\mbox{Jiazhen Pan}}
\author[1,13,14]{\mbox{Daniel Rueckert}}
\author[4]{\mbox{Krischan Braitsch\textsuperscript{$\ddagger$}}}
\author[2,3]{\mbox{Keno K. Bressem\textsuperscript{$\ddagger$}}}
\affil[1]{Chair for AI in Healthcare and Medicine, Technical University of Munich (TUM) and TUM University Hospital, Munich, Germany}
\affil[2]{Department of Diagnostic and Interventional Radiology, Klinikum rechts der Isar, TUM University Hospital, School of Medicine and Health, Technical University of Munich, Munich, Germany}
\affil[3]{Department of Cardiovascular Radiology and Nuclear Medicine, German Heart Center, TUM University Hospital, School of Medicine and Health, Technical University of Munich, Munich, Germany}
\affil[4]{Department of Medicine III, Klinikum rechts der Isar, TUM University Hospital, School of Medicine and Health, Technical University of Munich, Munich, Germany}
\affil[5]{Department of Radiology, Charit\'e -- Universit\"atsmedizin Berlin, Berlin, Germany}
\affil[6]{Department of Gastroenterology, Infectious Diseases and Rheumatology, Charit\'e -- Universit\"atsmedizin Berlin, Berlin, Germany}
\affil[7]{Chair of Medical Informatics, Institute of AI in Medicine and Healthcare, TUM School of Medicine and Health, Technical University of Munich, Munich, Germany}
\affil[8]{Clinical Department of Gynecology, TUM University Hospital, TUM School of Medicine and Health, Munich, Germany}
\affil[9]{TranslaTUM, Center for Translational Cancer Research, Technical University of Munich, Munich, Germany}
\affil[10]{Deutsches Konsortium f\"ur Translationale Krebsforschung, Heidelberg, Germany}
\affil[11]{Bavarian Cancer Research Center, Munich, Germany}
\affil[12]{Department of Engineering Science, University of Oxford, Oxford, UK}
\affil[13]{Munich Center for Machine Learning (MCML), Munich, Germany}
\affil[14]{Department of Computing, Imperial College London, London, UK}

\date{}  
\maketitle
\vspace{4em}
\noindent\textsuperscript{$\dagger$}\,These authors contributed equally to this work.\\
\noindent\textsuperscript{$\ddagger$}\,These authors share senior authorship.

\begin{abstract}
\noindent\textbf{Background.}
Multiple myeloma is managed through sequential lines of therapy over years to
decades, with each treatment decision depending on cumulative disease history
distributed across dozens to hundreds of heterogeneous clinical documents.
Whether large language model based systems can synthesise this evidence at a
level approaching expert agreement has not been established.

\noindent\textbf{Methods.}
A retrospective evaluation was conducted on longitudinal clinical records of
811 patients with multiple myeloma treated at a tertiary medical centre
between 2001 and 2026, covering 44{,}962 documents and 1{,}334{,}677
laboratory values, with external validation on MIMIC-IV. An agentic reasoning
system was compared against single-pass retrieval-augmented generation (RAG), iterative RAG, and full-context input on 469
patient--question pairs derived from 48 templates stratified into three
complexity levels. The reference standard was established by independent
double annotation from four oncologists with adjudication by a senior
haematologist.

\noindent\textbf{Findings.}
Iterative retrieval-augmented generation and full-context input converged on
a shared performance ceiling (75\D4\% versus 75\D8\%, Bonferroni-corrected
$p$ = 1\D00). The agentic system reached 79\D6\% concordance
(95\%\,CI 76\D4--82\D8), significantly exceeding both baselines
($+$3\D8 and $+$4\D2 percentage points; $p$ = 0\D006 and 0\D007). Gains
increased with question complexity, reaching $+$9\D4 percentage points on
criteria-based synthesis ($p$ = 0\D032), and with record length, reaching
$+$13\D5 percentage points in the top decile (exploratory, $n$ = 10).
The system error rate (12\D2\%) was comparable to expert disagreement
(13\D6\%), but severity distributions were inverted, with 57\D8\% of system
errors classified as clinically significant against 18\D8\% of expert
disagreements.

\noindent\textbf{Interpretation.}
Agentic reasoning was the only approach to exceed the shared performance
ceiling, with gains concentrated on the most complex questions and longest
records. The greater clinical consequence of residual system errors relative
to expert disagreement indicates that prospective evaluation in routine care
will be required before these findings translate into measurable patient
benefit.

\noindent\textbf{Funding.}
Bayern Innovativ (Bavarian State Ministry of Economics), Grant Number:
LSM-2403-0006.
\end{abstract}



\section*{Introduction}
\noindent Multiple myeloma is managed through sequential lines of therapy over years to decades, with each treatment decision depending on a cumulative record of prior exposures, documented responses, and evolving comorbidities that no single source can fully reflect~\cite{rajkumar2022multiple}. A typical patient accumulates dozens to hundreds of clinical documents over this trajectory, spanning laboratory data, pathology findings, imaging reports, and free-text notes, and the correct interpretation of any individual finding depends on temporal relationships that may span years~\cite{cui2025timer}. Determining whether a patient has progressed on a given regimen, whether prior toxicities preclude a planned therapy, or whether organ function meets eligibility criteria requires synthesis across document types and timepoints that no individual document can resolve. These demands fall on haematologists who already carry among the highest electronic health record burdens of any clinical specialty, with recent data showing more than 575 minutes per week in the record and over 219 minutes of after-hours documentation~\cite{holmgren2025oncology}, a burden that scales with the complexity of the disease trajectory.\\

\noindent Large language models (LLMs) have been applied to clinical record navigation across a range of tasks~\cite{du2025performance}. Published approaches fall into three broad strategies: single-pass retrieval-augmented generation (RAG), iterative RAG, and agentic systems that reason over specialised tools~\cite{liu2025improving, qiu2024llm, truhn2026artificial}. Earlier work on guideline interpretation, domain-tuned question answering, and factual verification showed that retrieval can reduce hallucination on well-defined single-document tasks~\cite{ferber2024gpt, lu2025enhancing, liu2025verifact, chung2025verifying}. More recently, Myers and colleagues tested longitudinal reasoning directly, comparing RAG against full-record long-context input on three tasks over hospitalised patients, finding equivalent performance at 128{,}000 tokens of context capacity and concluding that the absence of expert-adjudicated longitudinal datasets had prevented any test of whether this convergence holds for the harder multi-source synthesis tasks that dominate real clinical practice~\cite{myers2025evaluating}.\\

\noindent Agentic approaches have been evaluated primarily in two settings: diagnostic dialogue agents operating on simulated patients~\cite{tu2025towards, schmidgall2024agentclinic}, and tool-using agents tested against structured databases where each task can be resolved from a small number of entries~\cite{shi2024ehragent, jiang2025medagentbench, lee2025fhir}. Neither setting captures the demand of answering a treatment planning question from multiple instances of unstructured, partially contradictory longitudinal documentation. No evaluation of any such approach has been reported for multiple myeloma or any comparable disease trajectory, and whether clinically reliable performance is achievable against an independently annotated reference standard has not been established.\\

\noindent Here, a retrospective evaluation of expert concordance was conducted on institutional records of 811 patients with multiple myeloma treated at a tertiary medical centre over 25 years. An agentic reasoning system was compared against single-pass RAG, iterative RAG, and full-context input, all using the same locally deployed open-weight LLM to keep data within institutional infrastructure. The reference standard was established through independent double annotation by four oncologists with adjudication by a senior haematologist, residual inter-rater variability was explicitly classified, and the clinical safety profile was characterised through structured error severity analysis. External validity was assessed on the MIMIC-IV critical care database~\cite{PhysioNet-mimic-iv-note-2.2,johnson2023mimic}.\\

\section*{Methods}
\noindent The study was approved by the Ethics Committee of the Technical University of Munich (approval 2024-590-S-CB), and informed consent was waived given the retrospective design. Use of MIMIC-IV data was conducted under the PhysioNet credentialed data use agreement~\cite{goldberger2000physiobank}. All procedures followed the Declaration of Helsinki and applicable institutional guidelines.

\subsection*{Study design and data sources}
\noindent A retrospective evaluation of expert concordance was conducted on longitudinal multiple myeloma records from two independent institutions (Figure~\ref{fig:data}a). The primary dataset comprised all patients with a confirmed multiple myeloma diagnosis treated at TUM University Hospital between 1 January 2001 and 1 January 2026 (n=811). Textual reports were extracted from 44,962 German-language clinical documents, yielding a median of 55 documents per patient (IQR 20 to 76, Figure~\ref{fig:data}b) and a median follow-up of 6\D3 years from diagnosis (IQR 1\D7 to 9\D7, Figure~\ref{fig:data}c), with details of the extraction and conversion pipeline provided in Supplementary Methods~\ref{sup:pipeline}. Structured laboratory data (1,334,677 values) were extracted from the institutional laboratory information system and normalised to 731 canonical concepts, preprocessing details are provided in Supplementary Methods~\ref{sup:lab-data}. The external validation dataset comprised 716 patients identified by ICD code from the publicly available de-identified MIMIC-IV critical care database~\cite{johnson2023mimic}, yielding 26,767 clinical documents with a median of 26 documents per patient (IQR 11 to 53). This cohort was used without adaptation to assess transferability across institutions.

\subsection*{Clinical question bank and evaluation cohorts}
\noindent A bank of 48 clinical question templates covering core decision tasks in multiple myeloma management was developed in consultation with staff haematologists and reviewed for clinical representativeness. Templates were stratified into three complexity levels: single-record lookup (Level~1), temporal reasoning across multiple sources (Level~2), and criteria-based synthesis across document types and multiple timepoints (Level~3). The full template list with answer formats and scoring methods is provided in Supplementary Table~\ref{tab:question-bank}.

\noindent Three non-overlapping patient sets were defined prior to system development (Figure~\ref{fig:data}d, Figure~\ref{fig:consort-flow}). A development set of ten patients was used exclusively for iterative system development and is excluded from all reported evaluations. For the primary analysis, 100 patients were sampled from the remaining TUM cohort using stratified sampling over number of reports and recency of the last available data point, with each patient assigned five questions (two Level~1, two Level~2, one Level~3), yielding 500 patient-question pairs. For external validation, 20 patients were selected from the MIMIC-IV cohort using the same procedure, yielding 100 pairs.

\subsection*{Expert annotation and adjudication}
\noindent All pairs were annotated by independent chart review prior to system evaluation, with raters blinded to each other and to all system outputs. Each TUM patient was reviewed by exactly two of four oncologists with multiple myeloma expertise (AP, CN, JS, LM). Pairs with direct agreement were included without adjudication, disagreements were reviewed by a senior haematologist (KB) and classified into one of five categories. Full protocol details and operational definitions are provided in Supplementary Methods.

\noindent For the TUM cohort, direct agreement was reached on 65\D2\% of pairs, a further 28\D6\% were included after adjudication, and 6\D2\% were excluded, yielding 469 evaluable pairs (200 Level~1, 179 Level~2, 90 Level~3; Figure~\ref{fig:data}e). Pre-adjudication inter-rater agreement declined with complexity: $\kappa$ = 0\D69 at Level~1, $\kappa$ = 0\D60 at Level~2, and $\kappa$ = 0\D57 at Level~3 (Figure~\ref{fig:data}f). Among adjudicated cases, the most frequent categories were interchangeable or equivalent responses (39\D2\%) and clinically insignificant disagreement (36\D4\%), with clinically significant disagreement accounting for 8\D4\% (Figure~\ref{fig:data}g). For the MIMIC-IV cohort, 89 evaluable pairs were retained after the same procedure.

\subsection*{Agentic system and comparators}
\noindent The agentic system was designed to answer clinical questions requiring synthesis across temporally distributed, heterogeneous documentation. The system was distinguished from retrieval-based approaches by four architectural components (Figure~\ref{fig:workflow}). Task-relevant modules were selected from an indexed clinical skill library encoding question-type-specific reasoning protocols. An ordered tool-use plan with explicit stopping conditions was constructed prior to retrieval. A structured memory state encoding the user query, retrieved evidence, missing information, and stopping conditions was updated iteratively after each step. Iterative execution was carried out against purpose-built tools, including report and laboratory value retrieval with type and date filters and deterministic clinical scoring calculators, with automatic re-query on insufficient evidence. Final answers followed a schema-defined format with inline source citations. Full implementation details are provided in Supplementary Methods.

\noindent The same locally deployed 120-billion-parameter open-weight language model (gpt-oss-120b~\cite{agarwal2025gpt}) was used across all three comparator approaches and the same patient record database. In Simple RAG, single-pass dense retrieval was implemented without query rewriting or reranking~\cite{lewis2020retrieval}. Iterative RAG was extended to include subquery rewriting, hybrid BM25 and dense retrieval fusion, cross-encoder reranking, and a multi-round sufficiency loop~\cite{liu2025improving, doan2024hybrid}. In the Full Context configuration, retrieval was bypassed entirely, all documents and laboratory values were concatenated in reverse chronological order until the context window was filled~\cite{myers2025evaluating}. Two additional comparators are reported in Supplementary Table~\ref{tab:pairwise-bootstrap}, and detailed specifications for all approaches are provided in Supplementary Methods.

\subsection*{Error classification and citation sufficiency}
\noindent All patient-question pairs for which the agentic system diverged from expert consensus were classified by a senior haematologist (KB) into one of six categories: clinically significant error, clinically insignificant error, partially correct, acceptable or ambiguous, annotation error, and pipeline failure. The error classification taxonomy was aligned with the adjudication categories applied during expert annotation, enabling direct comparison of system error rates with expert disagreement rates at equivalent severity levels. To assess the reproducibility of this classification, a blinded inter-rater reliability sub-study was conducted by an independent oncologist (JS) on a proportional stratified random sample of 46 of the 115 divergent patient-question pairs, with categories additionally collapsed into three severity strata for analysis. Citation sufficiency was assessed by two reviewers (KB, CN) on a stratified sample of 96 responses, with each of the 48 question templates represented at least once. Full classification criteria, sampling details, and inter-rater reliability protocol are provided in Supplementary Methods.

\subsection*{Statistical analysis}
\noindent The primary outcome was concordance with expert consensus, defined as the proportion of pairs for which system output matched the adjudicated reference on substantive content. Single-value categorical items were scored as binary, whereas list-type items were scored using entry-level F1 computed against the reference list, contributing a continuous value between zero and one. Secondary outcomes were the clinically significant error rate derived from structured error classification and citation sufficiency. Each system was evaluated across ten independent runs to control for stochastic variability, and per-question scores were averaged across runs before analysis. Individual concordance estimates and their 95\% confidence intervals were computed by pair-level percentile bootstrap ($N_{\text{boot}} = 10{,}000$). Pairwise significance tests were computed via cluster bootstrap ($N_{\text{boot}} = 10{,}000$, $\alpha$ = 0\D05), resampling whole patients with replacement to account for within-patient correlation. Bonferroni correction was applied within each stratum independently, and all reported $p$-values are Bonferroni-corrected unless explicitly stated otherwise. The relationship between patient record length and system performance was examined as a hypothesis-driven exploratory analysis, with patients stratified into four quantile-based bins prior to analysis (Supplementary Table~\ref{tab:context-length-bins}).

\subsection*{Role of the funding source}
\noindent The funders of the study had no role in study design, data collection, data analysis, data interpretation, or writing of the report.
\section*{Results}
\subsection*{Accuracy against expert consensus}

\noindent The primary evaluation comprised 469 patient-question pairs across 100 patients from the TUM cohort, including 200 Level~1, 179 Level~2, and 90 Level~3 pairs retained after adjudication. Iterative RAG and Full Context performed equivalently, reaching 75\D4\% (95\%~CI [71\D9--78\D6]) and 75\D8\% ([72\D3--79\D3]) concordance with expert consensus respectively, with no statistically significant difference between them ($-$0\D4 percentage points, Bonferroni-corrected $p$~=~1\D00). Simple RAG trailed at 71\D5\% ([67\D8--75\D1]).

\noindent The agentic system achieved an overall concordance of 79\D6\% ([76\D4--82\D8]) (Figure~\ref{fig:results}a, Supplementary Table~\ref{tab:pairwise-bootstrap}), significantly higher than both Full Context ($+$3\D8 percentage points, $p$ = 0\D006) and Iterative RAG ($+$4\D2 percentage points, $p$ = 0\D007). Across ten independent evaluation runs, the agentic system showed a standard deviation of 1\D1~percentage points (Supplementary Table~\ref{tab:stability}). The skill library was identified by ablation analyses as the principal driver of performance. Its removal reduced overall concordance by 3\D0 percentage points to 76\D6\%, whereas removal of type and date filters in retrieval tools, deterministic clinical scoring tools, structured memory state, or pre-planned tool use individually reduced concordance by at most 0\D4 percentage points (Supplementary Table~\ref{tab:ablations}).

\noindent External validation on 89 evaluable patient-question pairs across 20 patients from the MIMIC-IV cohort preserved the system ranking (Figure~\ref{fig:results}e). The agentic system achieved an overall concordance of 84\D9\% ([77\D8--91\D2]), compared with 79\D2\% ([71\D4--86\D3]) for Simple RAG, 77\D9\% ([70\D1--85\D3]) for Iterative RAG, and 74\D1\% ([65\D5--82\D2]) for Full Context. The three non-agentic approaches clustered within overlapping confidence intervals, and the agentic system retained the highest concordance despite a change in documentation language (English versus German), institutional conventions, record structure, and the de-identification date-shifting applied to the source data. No system adaptation was performed between cohorts.

\subsection*{Performance across question complexity and record length}

\noindent Concordance decreased with question complexity across all approaches, but not uniformly (Figure~\ref{fig:results}a). For the agentic system, concordance fell from 86\D1\% ([81\D9--90\D1]) at Level~1 to 79\D5\% ([74\D1--84\D5]) at Level~2 and 65\D1\% ([56\D8--73\D3]) at Level~3. The advantage of the agentic system over Full Context increased monotonically with complexity, from $+$1\D0 percentage point at Level~1 ($p$ = 1\D00) to $+$3\D9 percentage points at Level~2 ($p$ = 0\D25) and $+$9\D4 percentage points at Level~3 ($p$ = 0\D032) (Figure~\ref{fig:results}b). A similar gradient was observed relative to Iterative RAG, with a difference of $+$9\D7 percentage points at Level~3 ($p$ = 0\D049) but no significant difference at Levels~1 or~2. At Level~3, the three non-agentic approaches ranged from 48\D9\% to 55\D7\%, whereas the agentic system reached 65\D1\%.

\noindent Concordance was stratified by total patient record length in a hypothesis-driven exploratory analysis (Figure~\ref{fig:results}c). Patients were divided into four bins: three terciles of the lower 90th percentile ($\leq$127k, 127 to 282k, and 282 to 541k~characters) and the top decile ($>$541k characters, corresponding to approximately 245{,}000 tokens; $n = 10$ patients, 47 evaluable pairs). In the lower three bins, no statistically significant differences were observed between the agentic system and either Full Context or Iterative RAG. In the top decile ($n$ = 10 patients, 47 evaluable pairs; exploratory), the agentic system achieved 75\D3\% concordance compared with 63\D9\% for Iterative RAG ($+$11\D4 percentage points) and 61\D8\% for Full Context ($+$13\D5 percentage points), bootstrap confidence intervals for these comparisons were wide owing to the small patient count and these findings should be interpreted as hypothesis-generating. From the shortest to the longest bin, concordance fell by 7\D1 percentage points for the agentic system, compared with 16\D9 for Iterative RAG, 19\D6 for Full Context, and 26\D5 for Simple RAG.

\subsection*{Clinical error profile and failure mechanisms}
\noindent Error classification was performed on a single evaluation run with overall concordance of 77\D8\%. All 115 patient-question pairs for which the agentic system diverged from expert consensus on that run, including list-type responses with partial but incomplete overlap (F1 between zero and one), underwent structured classification by a senior haematologist into one of six categories (Table~\ref{tab:error-classification}). Thirty-three divergences (28\D7\%) were classified as clinically significant errors, 24 (20\D9\%) as clinically insignificant errors, six (5\D2\%) as partially correct, 39 (33\D9\%) as acceptable or ambiguous responses in which both the system output and the reference annotation were defensible from the record, 11~(9\D6\%) as annotation errors in which the system response was correct and the reference annotation was not, and two (1\D7\%) as pipeline failures. The clinically significant error rate was 33 of 469 pairs (7\D0\%), with clinically significant errors distributed across complexity levels: 14 at Level~1, 12 at Level~2, and seven at Level~3. Blinded re-annotation of a proportional stratified random sample of 46 divergent pairs by an independent rater yielded a six-category $\kappa$ of 0\D667 ($95\,\%$ CI [0\D510, 0\D833]) and a three-stratum $\kappa$ of 0\D802 ([0\D687, 0\D962]), indicating substantial to near-perfect reproducibility of the classification across severity levels.

\noindent The system error rate, combining clinically significant and clinically insignificant categories, was 57 of 469 pairs (12\D2\%), comparable to the rate of expert disagreement on the same tasks (64 of 469 pairs, 13\D6\%, Figure~\ref{fig:data}g). Among system errors, 57\D8\% (33 of 57) were clinically significant, and among expert disagreements, 18\D8\% (12 of 64) were clinically significant. The severity distributions were therefore inverted despite comparable overall rates, with system errors carrying greater average clinical consequence than expert disagreements on the same tasks.

\noindent Citation sufficiency was assessed on a stratified sample of 96 agentic system responses from the same run, drawn evenly across complexity levels and concordance status, with each of the 48 question templates represented at least once (Figure~\ref{fig:results}d). All concordant responses were rated as fully supported by the retrieved source documents across all three complexity levels. Among discordant responses, the proportion rated as fully supported was 50\% at Level~1, 62\% at Level~2, and 50\% at Level~3. Of the 22 responses with citation insufficiency, incomplete retrieval was the dominant failure mechanism, accounting for 83\% of fully unsupported responses ($n = 18$), in the remaining 17\%, a relevant document had been retrieved but was not incorporated into the final response.
\section*{Discussion}
\noindent Retrieval-augmented and full-context approaches converged on a shared performance ceiling on real longitudinal myeloma records, and agentic reasoning was the only approach to exceed it. The advantage was not uniform. It was negligible on simple single-record lookups, grew monotonically with question complexity, and was largest for patients with the longest and most complex disease trajectories, precisely those for whom the clinical stakes of accurate synthesis are highest.

\noindent The convergence of Iterative RAG and Full Context at a shared ceiling indicates that the limiting factor is not the quantity of evidence presented to the model but the structure imposed on its integration. RAG delegates decomposition, evidence weighting, and clinical decision rules to a single generation step. Agentic reasoning externalises these operations as an explicit planning phase, iterative evidence gathering with re-query on insufficient results, and deterministic scoring calculators where clinical rules can be encoded without relying solely on language model interpretation. Ablation analyses indicate that the skill library contributes the largest individually separable gain, reducing concordance by 3\D0 percentage points when removed, while the remaining components each contribute at most 0\D4~percentage points in isolation, consistent with their role as interdependent scaffolding through which question-type-specific reasoning protocols are executed. The present findings extend the convergence ceiling previously observed on simpler clinical question answering~\cite{myers2025evaluating} to the harder longitudinal synthesis setting and show that structured reasoning is the first approach demonstrated to exceed it.

\noindent The clinical profile of the agentic advantage identifies a coherent deployment target. The gap over the strongest non-agentic approach widened with both question complexity and total record length, concentrating on patients with the most treatment lines, relapse-remission cycles, and accumulating eligibility constraints. A decade-long myeloma trajectory with multiple sequential regimens and evolving comorbidities is simultaneously the case where manual chart review consumes the most physician time and the case where accurate longitudinal synthesis is most decision-relevant. Performance differences were negligible for patients in the lower 90th percentile of record length and largest in the top decile (exploratory, $n$ = 10 patients), and the agentic advantage over Full Context increased from 1\D1 percentage points at Level~1 to 9\D4 percentage points at Level~3.

\noindent The system error rate (12\D2\%) was comparable to the rate of expert disagreement on the same tasks (13\D6\%), but the severity distributions were inverted: among system errors, 57\D8\% were clinically significant, against 18\D8\% of expert disagreements. The system therefore operated within the bounds of human annotator variability in aggregate, while its errors carried greater average clinical consequence. This distinction is not captured by concordance metrics alone and defines the gap between matching expert agreement in aggregate and readiness for clinical deployment. One measurement asymmetry should be noted, expert disagreements were measured before adjudication, whereas system divergences were measured against the finalised reference standard.

\noindent The dominant failure mechanism differed with task complexity. Errors on single-record lookup tasks arose predominantly from incomplete retrieval, as relevant documents were not returned by the search tools, a pattern directly addressable through retrieval engineering. Errors on multi-criterion synthesis tasks arose when the relevant evidence had been retrieved but was not correctly integrated during reasoning, a pattern consistent with the documented tendency of long-context language models to underweight information positioned in the middle of their input~\cite{liu2024lost}. The same mechanism likely explains part of the Full Context underperformance, where relevant passages were present in the context window but were not reliably attended to. Both limitations are properties of the underlying language model rather than of the agentic architecture. The model used here (gpt-oss-120b) was selected to run reproducibly on a single H200 GPU within institutional infrastructure, and larger or more recent models with stronger long-context reasoning may narrow the gap attributable to retrieval and integration failures.

\noindent External validation on the MIMIC-IV critical care database preserved both the system ranking and the complexity gradient despite a change in documentation language, institutional conventions, and record structure, with no system adaptation between cohorts. The external cohort is small, and confirmation on larger prospectively defined datasets across further institutions remains needed.

\subsection*{Limitations}
\noindent The question bank was developed with haematologists at the same institution and finalised before system development, so implicit alignment between question design and system architecture cannot be excluded. The senior haematologist who drafted the question bank also served as sole adjudicator of annotation disagreements without independent second review, though error classification by the same rater showed substantial to near-perfect reproducibility on blinded re-annotation of a stratified random sample ($\kappa$ = 0\D667 for six categories, 0\D802 for three strata). Error classification was performed without formal blinding to system performance, introducing potential bias in clinical significance assignments. Expert disagreements were classified before adjudication whereas system divergences were measured against the finalised reference standard, an asymmetry that may favour the system. The agentic advantage was consistent across three of five backbone models but attenuated for the two with stronger baseline performance, suggesting its magnitude depends on the long-context integration capabilities of the underlying model (Supplementary Table B.6). Comparison with proprietary frontier models was precluded by institutional data privacy requirements. The evaluation is retrospective, and whether agentic assistance improves clinical decisions in prospective use has not been assessed.

\subsection*{Conclusion}
 
\noindent Agentic reasoning answered clinical questions on years of real longitudinal myeloma records at a level of concordance with expert consensus that retrieval-augmented and full-context approaches did not reach, and its advantage increased with the complexity of cases. The total system error rate fell within the range of pre-adjudication inter-expert variability on the same tasks, although residual system errors carried greater average clinical consequence. Prospective evaluation in routine clinical care, with treating clinicians interacting with system outputs, will be required to determine whether these findings translate into measurable benefit at the point of care.

\subsection*{Contributors}
\noindent JM, LCA, KB, and KKB conceived and designed the study. JLu developed and coordinated the data acquisition and preprocessing pipeline, and contributed to system design and evaluation. JLa, SZ, MG, and MH contributed to data acquisition. CN, LM, AP, and JS contributed to the clinical question bank and annotated the data. KB drafted the clinical question bank, served as final adjudicator for all annotation disagreements, and contributed to the citation sufficiency and error classification reader studies. CN contributed to the error classification reader study. JM designed the agentic system, executed the experiments, performed result analysis and interpretation, and drafted the manuscript. CZ, ZB, FD, HH, AN, FP, and AZ provided iterative feedback on study design, execution, and interpretation, and revised the manuscript critically for important intellectual content. MM, FB, JP, DR, and KKB contributed to study design, supervised the study, and revised the manuscript. JM, JLu, KB, and KKB directly accessed and verified the underlying patient-level data reported in this study. All authors reviewed and approved the final version of the manuscript and accept responsibility for the decision to submit for publication.

\subsection*{Declaration of interests}
\noindent AP received travel support from Janssen-Cilag and Kite Gilead. FB received honoraria from Amgen, Johnson \& Johnson, Bristol Myers Squibb (BMS), AbbVie, and GSK, travel support from Amgen, Johnson \& Johnson, and BMS, and served on advisory boards for Amgen, Johnson \& Johnson, BMS, AbbVie, and GSK. MH received honoraria from Johnson \& Johnson, Sanofi, GSK, Oncopeptides, and Pfizer, payment for expert testimony from Johnson \& Johnson and Oncopeptides, travel support from Johnson \& Johnson, Oncopeptides, Pfizer, and Amgen, and served on advisory boards for Johnson \& Johnson, Sanofi, Oncopeptides, and Pfizer. JLa received a Google Gemma Academic Program Award, received speaker honoraria from the Forum for Continuing Medical Education (Germany), AstraZeneca (Germany), and Novartis (Germany), and served on an advisory board for Novartis (Germany). All other authors declare no competing interests.

\subsection*{Data sharing}
\noindent The MIMIC-IV database is publicly available under the PhysioNet credentialled data use agreement at \url{https://doi.org/10.13026/1n74-ne17}. The in-house TUM dataset cannot be shared publicly owing to patient privacy regulations and institutional data governance requirements. Anonymised aggregate results supporting the findings of this study are available from the corresponding author (johannes.moll@tum.de) on reasonable request from the date of publication. Requests will be reviewed by the study team and a signed data access agreement will be required before release. 

\subsection*{Acknowledgments}
\noindent This study was financed by the public funder Bayern Innovativ (Bavarian State Ministry of Economics) Nuremberg, Grant Number: LSM-2403-0006. KKB and LCA are further grateful to be supported by the Else-Kr\"oner-Fresenius-Foundation (2024\_EKES.16, 2025\_EKES.03). FB receives funding by the German Research Foundation (DFG) (TRR 387/1 - 514894665) and DFG BA 2851/7-1 (project ID: 537477296). 
\bibliographystyle{vancouver}      
\bibliography{lancet-sample}          
\section*{Figures}
\subsection*{Figure 1}
\begin{figure}[H]
\centering
\includegraphics[width=0.9\columnwidth]{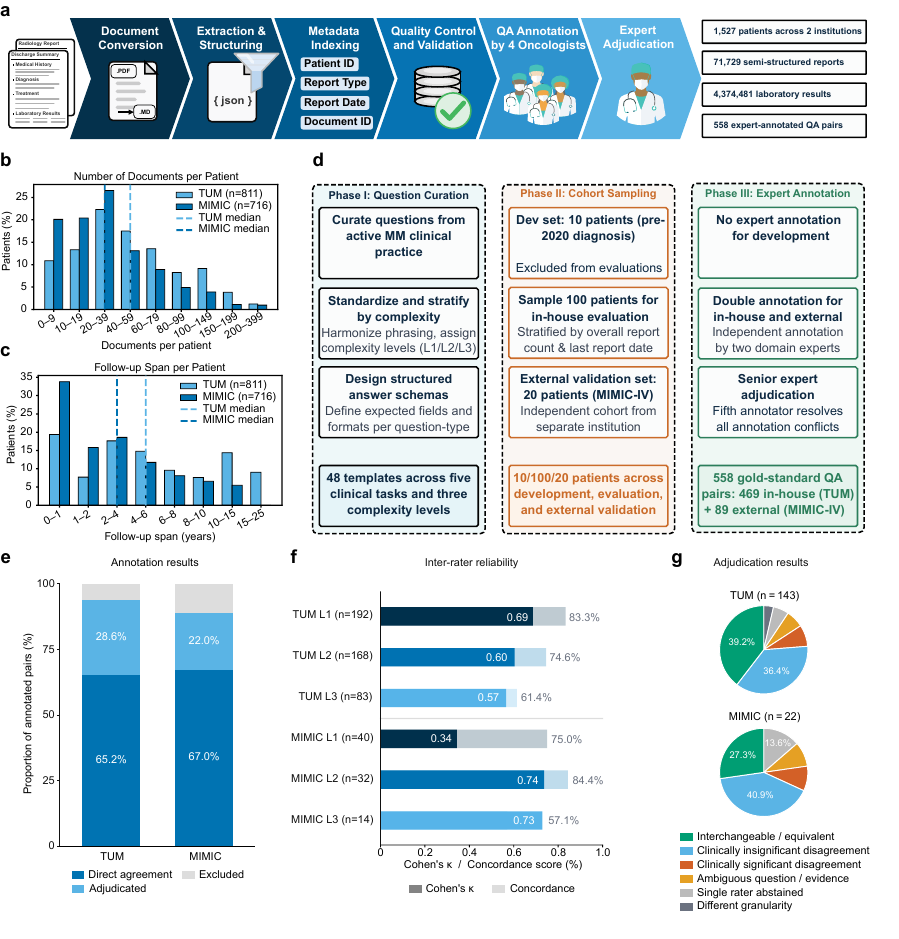}
\caption{
\textbf{Construction of longitudinal cohorts and expert-annotated evaluation dataset enabling clinically grounded assessment of longitudinal reasoning.}
(a) Overview of data sources and preprocessing pipeline across two institutions, including document extraction, structuring, metadata indexing, and quality control applied to heterogeneous clinical records.
(b) Distribution of document counts per patient, demonstrating substantial variability in record density and reflecting the complexity of real-world longitudinal documentation.
(c) Distribution of follow-up duration, highlighting long-term disease trajectories in the TUM cohort compared with shorter observation windows in MIMIC-IV.
(d) Study design and cohort construction, including development, in-house evaluation, and external validation sets.
(e) Annotation outcomes showing proportions of direct agreement, adjudicated cases, and exclusions.
(f) Inter-rater reliability across predefined complexity levels, reported as Cohen's $\kappa$ and observed agreement, illustrating moderate agreement for clinically complex tasks. The low $\kappa$ at MIMIC Level~1 reflects high prevalence of negative responses inflating the chance-agreement baseline.
(g) Distribution of adjudication categories, indicating that a substantial proportion of disagreements reflects clinically insignificant or interchangeable interpretations rather than true errors.
}
\label{fig:data}
\end{figure}

\newpage
\subsection*{Figure 2}
\begin{figure}[!h]
\centering

\includegraphics[width=\columnwidth]{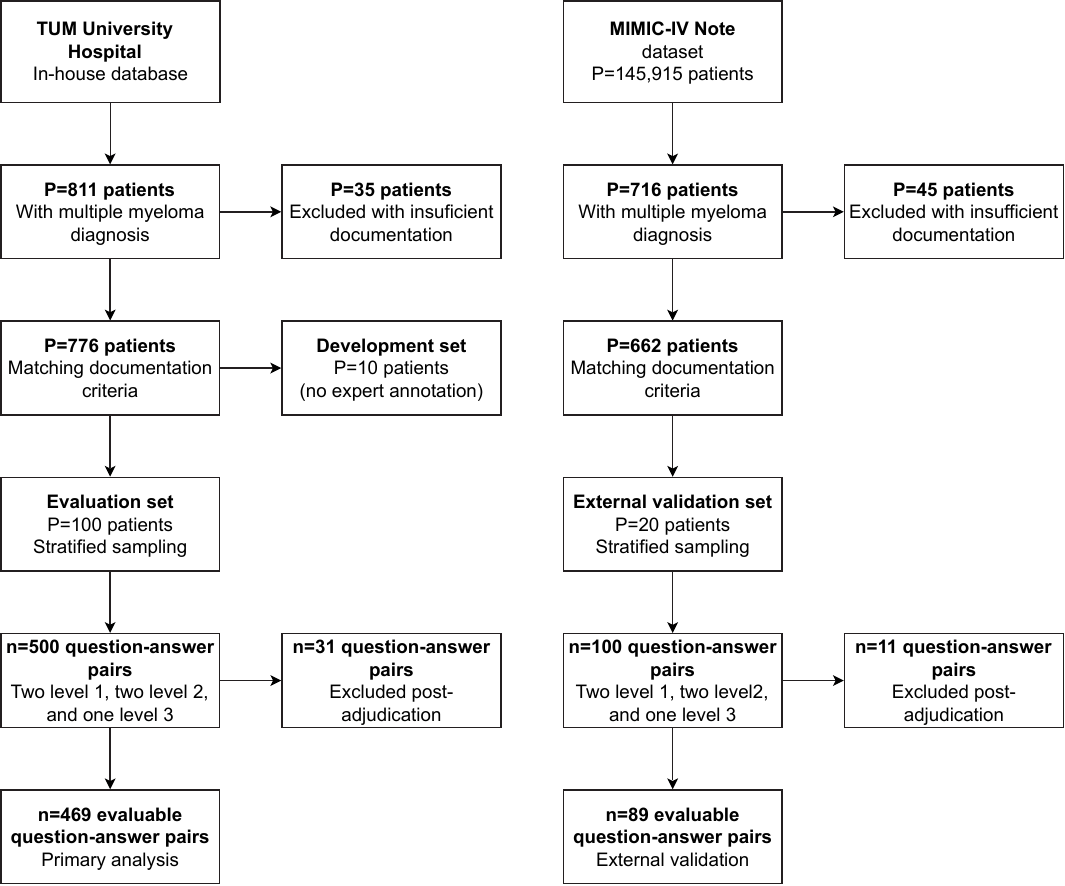}
\caption{
\textbf{Cohort selection yields representative evaluation sets for longitudinal clinical reasoning tasks.}
Flow diagram of patient inclusion from institutional (TUM) and external (MIMIC-IV) datasets, including filtering based on diagnosis and documentation criteria.
The primary evaluation cohort comprised 100 patients with 500 annotated question--answer pairs, of which 469 were retained after adjudication.
The external validation cohort included 20 patients with 100 annotated pairs, of which 89 were retained.
Stratified sampling ensured coverage across varying documentation density and temporal extent, supporting evaluation of reasoning under heterogeneous real-world conditions.
}
\label{fig:consort-flow}
\end{figure}

\newpage
\subsection*{Figure 3}
\begin{figure}[!h]
\centering

\includegraphics[width=\columnwidth]{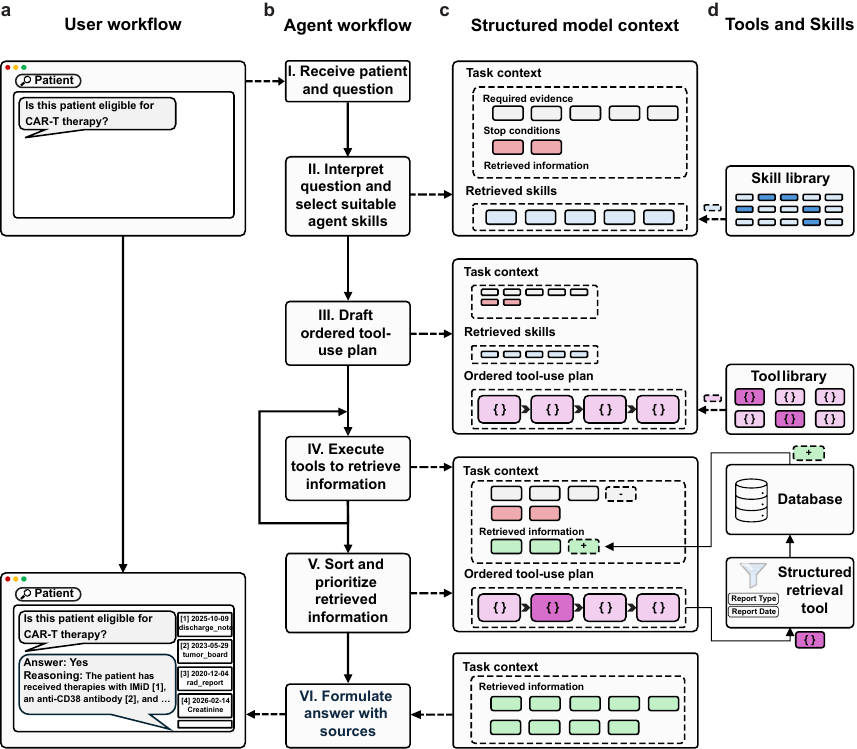}
\caption{
\textbf{Agentic system enables structured, traceable clinical reasoning across longitudinal patient records.}
(a) User-facing workflow illustrating query input and generation of citation-backed answers grounded in patient records.
(b) Internal agent workflow, including question interpretation, retrieval of task-specific clinical skills, generation of an ordered tool-use plan, iterative evidence retrieval, and synthesis.
(c) Structured model context integrating task requirements, retrieved domain knowledge, intermediate evidence, and stopping criteria for reasoning completion.
(d) Tool and skill library supporting structured access to clinical reports, laboratory trajectories, and deterministic scoring systems.
Together, these components enable multi-step reasoning in which each intermediate step is explicitly linked to retrieved evidence, ensuring that final answers are verifiable and traceable to source documents.
}\label{fig:workflow}
\end{figure}

\newpage
\subsection*{Figure 4}
\begin{figure}[!h]
\centering

\includegraphics[width=0.9\columnwidth]{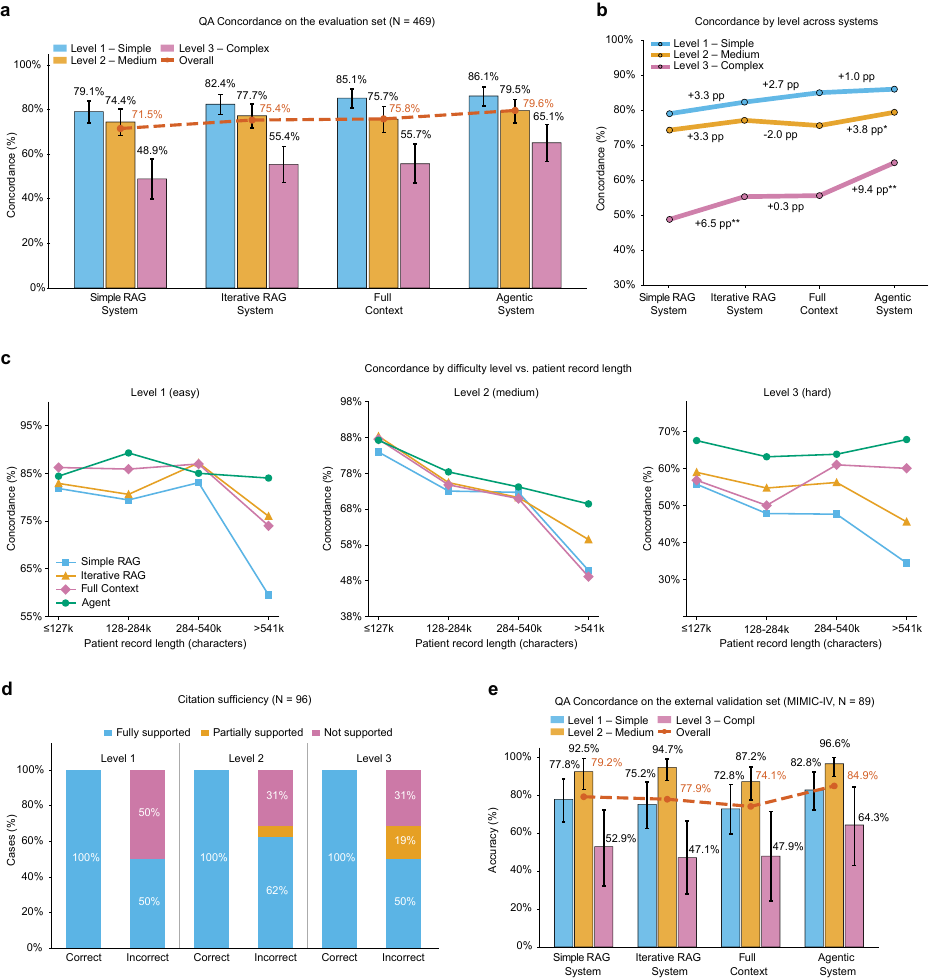}
\caption{
\textbf{Agentic reasoning improves accuracy, with largest gains in clinically complex tasks requiring longitudinal synthesis.}
(a) Overall and stratified concordance with expert consensus across system configurations on the primary evaluation cohort, showing superior performance of the agentic system.
(b) Performance differences by complexity level, demonstrating that the advantage of the agentic approach increases with task difficulty.
(c) Concordance stratified by patient record length and complexity level, showing convergence of all systems for shorter records and the largest agentic advantage among patients in the top decile of record length ($>$541k characters; exploratory, $n$ = 10 patients), where non-agentic configurations decline sharply.
(d) Citation sufficiency analysis showing the proportion of responses fully, partially, or not supported by retrieved source documents across complexity levels and concordance status.
(e) External validation on the MIMIC-IV cohort, confirming preservation of system ranking and robustness across institutions and documentation structures.
}
\label{fig:results}
\end{figure}
\newpage

\newpage
\subsection*{Table 1}
\begin{table}[H]
\centering
\small
\caption{\textbf{Error classification of agentic system divergences from expert consensus.} All 115 patient-question pairs for which the agentic system output diverged from the adjudicated reference annotation on a single evaluation run (overall concordance 77\D8\%) were classified by a senior haematologist. The clinically significant error rate was 33 of 469 evaluable pairs (7\D0\%).}
\vspace{6pt}
\label{tab:error-classification}

\begin{tabular}{lcccc}
\hline
\textbf{Classification} & \textbf{Overall} & \textbf{Level~1} & \textbf{Level~2} & \textbf{Level~3} \\
\hline
Clinically significant error    & 33 (28\D7\%) & 14 & 12 &  7 \\
Clinically insignificant error  & 24 (20\D9\%) &  5 & 11 &  8 \\
Partially correct               &  6 (5\D2\%)  &  1 &  5 &  0 \\
Acceptable / ambiguous          & 39 (33\D9\%) &  9 & 12 & 18 \\
Annotation error                & 11 (9\D6\%)  &  3 &  6 &  2 \\
Pipeline failure                &  2 (1\D7\%)  &  1 &  0 &  1 \\
\hline
Total                           & 115          & 33 & 46 & 36 \\
\hline
\end{tabular}
\end{table}

\newpage
\setcounter{section}{0}
\renewcommand{\thesection}{A}
\section{Supplementary Methods}

\subsection{Document preprocessing and extraction pipeline}
\label{sup:pipeline}
\noindent Discharge summaries, radiology reports, pathology reports, tumour board proceedings, and ancillary document types (cardiology, cytology, flow cytometry, and genomic diagnostics reports) were exported as PDF files from the institutional SAP electronic health record system at TUM University Hospital and converted to structured Markdown using a rule-based module that applied font, size, positioning, and layout metadata to exclude non-clinical content and to identify section boundaries, which were preserved as hierarchical headers in the output.

\noindent A post-processing step merged sections shorter than 50 words into the nearest adjacent section and split sections exceeding 350 words into overlapping chunks with a 50-word overlap. Laboratory reports stored as structured data were handled separately and excluded from text segmentation.

\noindent Document-level metadata (patient identifier, report type, report date, and document identifier) were captured at export and stored alongside each section as structured fields, enabling retrieval queries to filter by type, date, or patient without parsing free text. Report type labels were mapped to a controlled vocabulary of nine canonical categories used as filter arguments by the agentic system's retrieval tool. Sections were indexed using FTS5 full-text search with Porter stemming, metadata fields were stored as unindexed columns to support type- and date-based filtering without affecting relevance scoring.

\subsection{Laboratory data normalisation and concept mapping}
\label{sup:lab-data}

\noindent Laboratory values were extracted from the institutional SAP system as timestamped observations comprising a test name, a numeric or categorical result, the reporting unit, and the laboratory-defined reference range. Because test names in the source system reflect free-text entry conventions that evolved over more than two decades of clinical use, a two-stage normalisation pipeline was developed to enable reliable programmatic retrieval.

\noindent In the first stage, a catalogue of canonical laboratory concepts was constructed from all test names observed across the myeloma cohort. Each unique institutional code was assigned a stable identifier derived from a deterministic hash of its normalised form. Text normalisation comprised case folding, transliteration of German umlauts (\textit{\"a}~$\rightarrow$~\textit{ae}, \textit{\"u}~$\rightarrow$~\textit{ue}, \textit{\"o}~$\rightarrow$~\textit{oe}, \textit{\ss}~$\rightarrow$~\textit{ss}), removal of punctuation and special characters, and whitespace collapsing, yielding 731 unique canonical codes.

\noindent In the second stage, an alias index was built to resolve synonymous test names to canonical codes. A total of 1,054 observed name variants were mapped to 736 canonical target codes, producing 1,404 alias entries. Of the canonical codes, 228 (31\%) were associated with two or more synonymous input names, reflecting institutional naming changes over time, the coexistence of abbreviated and full-length designations (e.g.\ \textit{crp (c-reakt.\ pr)} and \textit{crp (c-reakt.\ protein)}), and specimen-type suffixes (e.g.\ \textit{albumin}, \textit{albumin sm}, \textit{albumin l}). A set of search terms was pre-computed per canonical code from tokenised variants to support fuzzy matching at query time.

\noindent The resulting catalogue spans myeloma-specific markers (serum and urine immunoglobulin free light chains $\kappa$ and $\lambda$, free light chain ratio, serum protein electrophoresis, immunofixation, immunoglobulins G, A, M, and D, $\beta_2$-microglobulin), prognostic parameters (lactate dehydrogenase, C-reactive protein, calcium, albumin, haemoglobin, creatinine), and the broader set of routine haematological, biochemical, and coagulation tests documented over each patient's care. No unit conversion was applied, original test name, reporting unit, and institutional reference range were returned alongside each numeric result. Laboratory values were indexed by their original timestamps with no temporal windowing imposed.

\subsection{Clinical question bank: development and instantiation}

\noindent The clinical question bank was developed by a senior haematologist (KB) in consultation with the annotating oncologists (AP, CN, JS, LM) prior to any system development or evaluation. Templates were drafted to cover core clinical decision tasks in the longitudinal management of multiple myeloma, including current and prior treatment status, treatment response assessment, toxicity and dose modification documentation, diagnostic workup completion, disease staging, comorbidity scoring, and therapy eligibility determination. The initial template set was revised through iterative consensus discussion until coverage of the intended task space was judged representative. All revisions were completed before the evaluation cohort was defined and before any system outputs were generated.

\noindent Each template was assigned by consensus to one of three complexity levels based on the cognitive operations required to produce a correct answer. Level~1 (single-record lookup) comprised questions answerable from a single document or structured data entry, such as whether the patient was currently receiving a specific medication. Level~2 (temporal reasoning) required integration of information across multiple documents or timepoints, such as identifying all documented treatment intervals for a given agent or determining the best documented response under a specified regimen. Level~3 (criteria-based synthesis) required structured synthesis across document types, laboratory values, and clinical timepoints against a multi-criterion decision rule, such as evaluating eligibility for BCMA-directed CAR-T cell therapy. The final bank comprised 48 templates (20 at Level~1, 18 at Level~2, and 10 at Level~3) spanning five clinical task categories (single choice, treatment intervals, first occurrence, staging, and eligibility). The full template list is provided in Supplementary Table~\ref{tab:question-bank}.

\noindent The answer schema for each template determined the scoring method. Templates requiring a single categorical response (a yes/no/not documented answer, a single date, or a single staging score) were assigned binary scoring. Templates requiring enumeration of multiple entries (cycle start dates with doses, treatment intervals) were assigned list-type F1 scoring, in which precision, recall, and their harmonic mean were computed over the set of response entries against the reference list.

\noindent For each patient in the evaluation cohort, five questions were instantiated from the template bank (two at Level~1, two at Level~2, and one at Level~3) by random draw without replacement from a shuffled deck at each complexity level (seed~=~42). When the deck was exhausted it was reshuffled, ensuring balanced reuse across patients. Date placeholders were replaced with the date of the last available clinical report for that patient. For the MIMIC-IV cohort, de-identified dates were used as recorded in the dataset, which applies a consistent per-patient shift that preserves temporal ordering and intervals. Whether a given question was answerable from a particular patient's record was determined only during expert annotation. The same instantiation procedure was applied to both cohorts.

\subsection{Annotation protocol}
\label{sup:annotation}

\noindent Prior to annotation, all four oncologists (AP, CN, JS, LM) received structured training from the senior haematologist (KB) on the question bank, the answer schema conventions, the complexity-level definitions, and the criteria for abstention. Each annotator was assigned 50 patients from the TUM evaluation cohort by randomised balanced allocation such that each patient was reviewed by exactly two annotators and no annotator pair was systematically overrepresented (seed~=~42). For the MIMIC-IV cohort, 20 patients were distributed between two of the same annotators using the same allocation procedure.

\noindent Annotation was performed using a purpose-built Streamlit web application that presented each question with schema-aware input widgets (single-choice dropdowns, multi-interval entry forms, criteria tables with per-criterion status fields, date entry, and free-text fields). Annotators were instructed to use the abstain checkbox when an answer was not clearly supported by available documentation up to the cutoff date, and could optionally select one or more evidence documents and provide free-text comments. All responses were persisted to CSV logs with timestamps.

\noindent For the TUM cohort, annotators reviewed patient records using the institutional SAP interface, which they use routinely in clinical practice. Each question specified a cutoff date, annotators were instructed to disregard documentation dated after the cutoff and, for conflicting statements, to prefer the most recent documentation before that date. For the MIMIC-IV cohort, annotators accessed an equivalent data browser providing report viewing and structured laboratory retrieval over the de-identified database. Annotators were blinded to each other's responses throughout, the application enforced single-user login and stored responses in separate log files. No model-generated content was displayed at any point during annotation.

\subsection{Adjudication rules and disagreement taxonomy}
\label{sup:adjudication}

\noindent All patient-question pairs for which two independent annotators did not reach direct agreement were referred to the senior haematologist (KB) for adjudication. Three case types entered adjudication: categorical disagreements (both annotators provided a response but on differing substantive content), single-rater abstentions (one annotator provided a response and the other abstained), and dual abstentions (both annotators abstained). Dual abstentions were excluded from the evaluable set without adjudication.

\noindent The adjudicator reviewed each disagreement using the same record interface and cutoff-date convention as the original annotators. Both annotators' responses were presented side by side, including evidence selections and comments, and the adjudicator provided a final answer using the same schema-aware interface. The adjudicator could adopt either annotator's response or provide an independent answer, in single-rater abstention cases, the non-abstaining rater's response was independently verified rather than accepted by default. Pairs for which the adjudicator also abstained were excluded from the evaluable set.

\noindent Each adjudicated disagreement was classified into one of five categories. \textit{Interchangeable or equivalent responses} covered cases in which both answers were clinically synonymous or differed only in phrasing, abbreviation, or formatting. \textit{Clinically insignificant disagreement} covered cases in which one annotator made an error or omission that would not alter a clinical decision. \textit{Clinically significant disagreement} covered cases in which the two annotators reached substantively different clinical conclusions with different management implications. \textit{Ambiguous question or evidence} covered cases in which the record did not contain sufficient or unambiguous information, or the instantiated question admitted more than one defensible interpretation. \textit{Difference in response granularity} covered cases in which both annotators were correct but reported at different levels of detail. Single-rater abstentions were adjudicated but not assigned a disagreement category. The adjudicated answer served as the reference standard for all concordance analyses, for pairs with direct agreement, the shared answer was used without adjudication.

\subsection{Expert annotation and adjudication outcomes}
\label{sup:annotation-outcomes}
\noindent For the TUM cohort, direct agreement was reached on 326 of 500 pairs (65\D2\%). A further 143 (28\D6\%) were included after adjudication and 31 (6\D2\%) were excluded (29 dual abstentions, 2 adjudicator abstentions), yielding 469 evaluable pairs in the primary analysis (200 Level~1, 179 Level~2, 90 Level~3; Figure~\ref{fig:data}e). Pre-adjudication inter-rater agreement was $\kappa$ = 0\D69 at Level~1 (observed agreement 83\D3\%, $n = 192$), $\kappa$ = 0\D60 at Level~2 (observed agreement 74\D6\%, $n = 168$), and $\kappa$ = 0\D57 at Level~3 (observed agreement 61\D4\%, $n = 83$) (Figure~\ref{fig:data}f). Among adjudicated cases, the most frequent disagreement categories were interchangeable or equivalent responses (39\D2\%) and clinically insignificant disagreement (36\D4\%), with the remainder attributable to clinically significant disagreement (8\D4\%), ambiguous question or evidence (7\D0\%), differences in response granularity (3\D5\%), and single-rater abstentions (5\D6\%) (Figure~\ref{fig:data}g).

\noindent For the MIMIC-IV cohort, direct agreement was reached on 67\D0\% of pairs. A further 22\D0\% were included after adjudication and 11\D0\% were excluded, yielding 89 evaluable pairs. Pre-adjudication inter-rater agreement was $\kappa$ = 0\D74 at Level~2 (observed agreement 81\D2\%, $n = 32$) and $\kappa$ = 0\D73 at Level~3 (observed agreement 78\D6\%, $n = 14$). The lower Level~1 kappa ($\kappa$ = 0\D34, observed agreement 75\D0\%, $n = 40$) reflects the high prevalence of negative responses in the MIMIC-IV records (72\D5\% of Level~1 answers were ``False''), which inflates the chance-agreement baseline and suppresses $\kappa$ even when observed agreement is high.

\subsection{Agentic system: implementation details}

\noindent The agentic system was implemented as a four-phase reasoning pipeline. Each phase corresponded to a structured prompt (termed a \textit{signature}) whose output was parsed as JSON and validated before progression to the next phase. The four phases were question assessment and skill selection, tool-use plan construction, iterative tool execution, and final answer synthesis. A schematic overview is provided in Figure~2.

\subsubsection*{Phase I: Question assessment and skill selection.}

\noindent Upon receipt of a clinical question, the system derived a short medical analysis identifying the clinical intent and key hypotheses, a list of required and currently missing information points, and a preliminary complexity assessment. In parallel, a subset of skill modules was selected from the indexed clinical skill library based on question content and available skill summaries. Heuristic keyword matching served as a fallback to ensure that question-type-specific style skills were activated even when the language model omitted them, for example, questions containing temporal keywords (\textit{aktuell}, \textit{derzeit}) triggered the current-status style skill, and questions referencing eligibility criteria triggered the eligibility style skill. A base structured-annotation style skill was always included to enforce the two-line output format.

\subsubsection*{Clinical skill library.}

\noindent The skill library comprised modular instruction packages in four categories. \textit{Workflow skills} encoded task-specific decomposition strategies and evidence priorities for clinical tasks such as therapy reconstruction, eligibility assessment, staging score calculation, and laboratory trend interpretation. \textit{Parsing skills} provided normalisation rules for institutional abbreviations, German-language date formats, and entity synonymy (e.g., mapping trade names to generic drug names). \textit{Style skills} defined answer structure templates for each question type, including required fields and formatting conventions. \textit{Policy skills} encoded deterministic precedence rules for resolving contradictory evidence, including temporal authority (most recent document takes precedence), plan-versus-administered therapy disambiguation (administered therapy overrides documented plans), and contradiction resolution heuristics.

\noindent Skills were indexed by identifier and category. The language model selected relevant skill identifiers from a summary prompt listing all available skills. Selected skills were rendered into two context blocks: a workflow context block (workflow, parsing, and knowledge skills) injected during tool execution, and a style context block (style skills) injected during final answer synthesis. Policy skills were attached deterministically based on the selected style and workflow skills rather than by model selection, ensuring consistent application of temporal authority and contradiction resolution rules regardless of model selection behaviour. Across the evaluation cohort, an average of 6.7 skill modules were selected per question, corresponding to approximately 8,000 tokens of skill context, compared with approximately 49,000 tokens required to load all 41 modules unconditionally. The contribution of the selection mechanism to accuracy and efficiency is reported in the ablation analysis (Table~\ref{tab:ablations}).

\subsubsection*{Fixed baseline prompt.}

\noindent All system configurations, including ablation conditions in which the skill library was disabled, shared a fixed baseline prompt of approximately 500 tokens. This prompt provided task framing (instructing the model to answer clinical questions from retrieved context only), the two-line output format specification, all answer schema definitions with worked examples covering single-choice, list, multi-interval, cycle-start-dose-list, and criteria-table formats, citation formatting rules using bracketed context identifiers, the distinction between ``Nicht dokumentiert'' (information absent from the record) and ``Nein'' (documented negative finding), and drug name normalisation mappings covering drug class to substance mappings (CD38 antibodies, immunomodulatory drugs, proteasome inhibitors, CAR-T, and bispecific T-cell engagers), common abbreviations (Dara, Btz, Len), and standard regimen names (VRd, KRd, Dara-Rd, DVRd). Clinical reasoning workflows, temporal reasoning policies, disease-specific reference knowledge (staging criteria, response categories), and structured retrieval strategies were provided exclusively by the skill library, ensuring that ablation comparisons isolated the contribution of clinical reasoning protocols from formatting, schema, and normalisation knowledge.

\subsubsection*{Phase II: Tool-use plan construction.}

\noindent A structured tool-use plan was drafted as a JSON object containing an ordered array of steps and a set of global stopping conditions. Each step specified a step number, a natural-language objective, the tool to be invoked, the tool arguments as a JSON object, a list of evidence requirements, and a conditional stopping rule. Up to three plan construction attempts were permitted, if the language model returned an invalid or empty plan, a repair prompt was appended and the plan was regenerated. If all three attempts failed, the run was terminated as a pipeline failure. In ablation conditions where planning was disabled, tools were selected reactively at each execution step without a pre-drafted plan.

\subsubsection*{Phase III: Iterative tool execution.}
\noindent Across execution rounds, the system maintained a structured memory state as a JSON object encoding the original query, accumulated evidence from prior tool calls, outstanding missing information, and the global stopping conditions defined in the tool-use plan. This state was updated after each round and passed forward to inform tool selection, re-query decisions, and termination.

\noindent The plan was executed step by step, with up to eight tool-use rounds permitted per question. At each round, the language model received the current plan step, accumulated evidence from prior rounds, the list of still-missing information, and the set of allowed tools, then selected one of three actions. It could invoke a specified tool, advance to the next plan step without a tool call, or terminate execution and proceed to answer synthesis. If the model chose to terminate before executing at least one tool call for the current step and further planned steps remained, a repair prompt required it to either execute the planned tool or explicitly skip the step.

\noindent Duplicate queries (identical tool name and query string) were blocked and the model was notified. A negative query cache tracked report retrieval queries that returned no results, preventing re-execution of identical unsuccessful queries. Tool results were cached within each run to avoid redundant calls with identical arguments. An automatic adjustment mechanism progressively broadened retrieval parameters after repeated empty results: after one failure, the number of returned results was doubled (up to a cap of 30), after two failures, the query was reduced to its keyword subset, after three failures, temporal scope restrictions were removed. A context budget of 120,000 tokens was enforced, if the estimated token count of accumulated context nodes exceeded this limit, retrieval was terminated early and the system proceeded to answer synthesis. Context nodes were deduplicated by section identifier before being passed to the final answer phase.

\subsubsection*{Tool specifications.}

\noindent Five tool classes were available to the agentic system. The \textit{report retrieval tool} performed keyword search over the document database using BM25 ranking (BM25Okapi from the rank\_bm25 library). The tool accepted a free-text query string, a report type filter selecting from nine canonical document categories, a temporal scope (all records, the most recent document, a specific date, or a date range), and corresponding date parameters. A report type synonym mapping normalised common German-language and abbreviated report type names to canonical values, and query strings were sanitised to remove answer-schema tokens that the language model occasionally included in retrieval queries.

\noindent The \textit{laboratory query tool} fetched structured laboratory values by canonical marker key directly from the laboratory database table, accepting a list of canonical lab keys with the same temporal scope options as the report retrieval tool. The set of available canonical keys was determined per patient at runtime and provided to the language model in the prompt to prevent queries for non-existent markers. A default limit of five results per marker was applied.

\noindent Four deterministic clinical scoring calculators were provided, each implemented as a stateless function with validated inputs and JSON output. The \textit{ISS} (International Staging System) assigned Stage~I when serum $\beta_2$-microglobulin was below 3\D5~mg/L and serum albumin was at least 3\D5~g/dL, Stage~III when $\beta_2$-microglobulin was 5\D5~mg/L or above, and Stage~II otherwise. The \textit{R-ISS} (Revised ISS) assigned Stage~I when ISS Stage~I criteria were met together with the absence of high-risk cytogenetic abnormalities (del(17p), t(4;14), or t(14;16)) and normal LDH, Stage~III when ISS Stage~III criteria were met together with either high-risk cytogenetics or elevated LDH, and Stage~II otherwise. The \textit{R2-ISS} (Second Revision ISS) incorporated additional risk factors including the LDH/ULN ratio and 1q gain/amplification status, yielding four stages (I--IV). The \textit{HCT-CI} (Hematopoietic Cell Transplantation Comorbidity Index) was computed from 17 boolean comorbidity flags with weights of 1 (arrhythmia, cardiac disease, inflammatory bowel disease, diabetes requiring medication, cerebrovascular disease, psychiatric disturbance, mild hepatic abnormality, obesity with BMI above 35, and persistent infection), 2 (rheumatologic disease, peptic ulcer, moderate-to-severe renal impairment, and moderate pulmonary disease), or 3 (prior solid tumour, heart valve disease, severe pulmonary disease, and moderate-to-severe hepatic disease), with risk groups defined as low (score 0), intermediate (score 1--2), and high (score $\geq 3$).

\subsubsection*{Phase IV: Final answer synthesis.}

\noindent After tool execution, accumulated context nodes were deduplicated and assigned sequential citation identifiers. The language model received the original question, patient context, a plan execution summary, the formatted context snippets with citation identifiers, the style context from selected skills, the response requirements derived from the question assessment, and the output of the policy engine where applicable. The policy engine was a deterministic module that ranked retrieved evidence items by temporal authority and recency, resolved contradictions between planned and administered therapies, and produced a structured resolution (select, abstain, or conflict) passed to the final answer signature.

\noindent The final answer was required to conform to a two-line format comprising an ``Answer:'' line containing the schema-aligned value and a ``Reasoning:'' line containing one to two sentences with inline citation references. If the initial output did not conform to this format, a repair prompt was appended and the answer was regenerated (up to two attempts). If no valid answer was produced after all attempts, the run was recorded as a pipeline failure. Citation identifiers appearing in the answer were validated against the set of available context nodes, and citations referencing non-existent sources were flagged as potentially hallucinated.

\subsubsection*{Decoding and inference settings.}

\noindent All language models were deployed locally on a single NVIDIA H200 GPU using a vLLM inference server with LiteLLM as the routing layer, providing an OpenAI-compatible API endpoint. The primary model (gpt-oss-120b) was served in mxfp4 precision. Decoding parameters were set to the values recommended by the respective model developers and held constant within each model's evaluation, no parameter optimisation was performed on the evaluation cohort. Run-to-run variability attributable to stochastic decoding was characterised through the ten-run stability analysis (Supplementary Table~4).

\subsubsection*{Pipeline failure definition.}

\noindent A pipeline failure was recorded when the system did not return a parseable, schema-conformant response within the permitted retry budget (up to three attempts for tool-use plan construction and up to two for final answer synthesis). In the classified error run, two of 469 pairs (0\D4\%) were classified as pipeline failures.

\subsection{Comparator configurations}

\noindent Simple RAG, Iterative RAG, and Full Context are described in the main text. All comparators used the same locally deployed 120-billion-parameter language model (gpt-oss-120b), the same patient record database, and the same structured answer prompt requiring the two-line format. Temperature was set to 0\D2 for all comparator configurations. Maximum output tokens were set to 512 for answer generation and 256 for query generation where applicable. The embedding model used for dense retrieval was distiluse-base-multilingual-cased-v2, a multilingual sentence transformer producing normalised embeddings for cosine similarity computation.

\noindent The LLM Baseline configuration received only the patient identifier, the question text, a reference date, and the answer schema, with no record access or retrieval context of any kind. The system prompt instructed the model to answer based solely on its parametric knowledge, establishing a floor for performance without record access.

\noindent Advanced RAG extended Simple RAG with query rewriting into up to eight focused retrieval queries, hybrid fusion of BM25 and dense scores (BM25 built over character trigrams with $k_1 = 1\D2$ and $b = 0\D75$, fusion weight $\alpha = 0\D5$), and cross-encoder reranking using bge-reranker-v2-m3. The top 20 documents after reranking were packed into the context window subject to the same 120,000-token budget. This configuration is reported in Supplementary Table~\ref{tab:pairwise-bootstrap} but is not included in the main comparator set.

\subsection{Scoring and concordance computation}

\noindent The concordance metric is defined in the main text. For single-value categorical items, scoring was binary: a score of 1 was assigned if the system output matched the adjudicated reference in substantive content and 0 otherwise. Matching was performed after normalisation of both the system output and the reference annotation, including extraction of the answer value from the two-line output format, case-insensitive comparison, whitespace trimming, and removal of formatting artefacts. Format deviations that did not alter substantive content were not penalised. Semantic equivalences were applied where clinically appropriate, for therapy-related questions, ``nie verabreicht'' (never administered) and ``nicht dokumentiert'' (not documented) were treated as equivalent when the patient record contained no evidence the therapy had ever been given.

\noindent For list-type items, F1 was computed over the set of response entries against the reference list, with each entry treated as an atomic unit. Precision was the fraction of system entries matching a reference entry, recall was the fraction of reference entries matched by a system entry, F1 was their harmonic mean. For list-type items preceded by a status field, the status was evaluated separately: an incorrect status (e.g., ``Nie verabreicht'' when reference entries existed) resulted in a score of 0 for the entire item regardless of entry-level matching.

\noindent Overall concordance was computed as the mean score across all evaluable patient-question pairs, where each pair contributed either a binary score (0 or 1) or a continuous F1 score (0 to 1) depending on the answer schema type. Per-question scores were first averaged across ten independent runs, and overall concordance was computed from these per-question means.

\noindent Confidence intervals and pairwise significance tests were computed using a cluster bootstrap procedure. Whole patients were resampled with replacement, retaining all questions per sampled patient to account for within-patient correlation; the concordance statistic was recomputed on each resample over $N_{\text{boot}} = 10{,}000$ iterations. The 2\D5th and 97\D5th percentiles of the bootstrap distribution defined the 95\% confidence interval (two-sided). The two-sided $p$-value for pairwise differences was derived as the proportion of bootstrap resamples in which the absolute deviation of the resampled difference from its mean equalled or exceeded the absolute observed difference (shift-to-null method). Bonferroni correction was applied within each stratum using $m = 6$ headline pairwise comparisons, capped at 1\D0. Results of all pairwise comparisons are reported in Supplementary Table~\ref{tab:pairwise-bootstrap}.

\subsection{Error classification protocol}

\noindent Error classification was performed by the senior haematologist (KB) on the evaluation run with the lowest overall concordance among the ten independent runs (77\D8\%, compared with a 10-run mean of 79\D6\%), selected to provide the most conservative basis for error rate estimation. All 115 patient-question pairs for which the system output diverged from the adjudicated expert consensus, including list-type responses with partial but incomplete overlap, were submitted for classification.

\noindent Classification was performed using a purpose-built review application that presented each case with the question text, answer schema, the adjudicated reference annotation, and the full agent response comprising the extracted answer, the reasoning chain with inline citation identifiers, and all retrieved source documents. Retrieved documents were separated into those cited in the agent's reasoning and those retrieved but not cited, each could be expanded to display the full report text with the cited passage highlighted. The reviewer additionally had access to the institutional SAP interface to verify claims against the primary clinical record when needed.

\noindent Each divergence was classified into exactly one of six categories. \textit{Annotation error} (category A) covered cases in which the system response was correct and the adjudicated reference was wrong or incomplete. \textit{Acceptable or ambiguous} (category B) covered cases in which both the system response and the reference were clinically defensible given the available evidence. \textit{Partially correct} (category C) covered cases in which the system captured the correct clinical concept but with incomplete or imprecise details. \textit{Clinically insignificant error} (category D) covered cases in which the system response was incorrect but the error would not alter a clinical decision. \textit{Clinically significant error} (category E) covered cases in which the system response was incorrect in a way that could alter a clinical decision. \textit{Pipeline failure} (category F) covered cases in which the system did not produce a valid, schema-conformant response within the permitted retry budget. The threshold separating clinically insignificant from clinically significant errors was whether a treating clinician relying on the system response rather than the correct answer would be expected to make a different management decision.

\noindent The clinically significant error rate reported in the main text (7.0\%, 33 of 469 evaluable pairs) was defined as the proportion of category E errors over all evaluable pairs. Because the classified run had below-average concordance, this rate represents a conservative upper bound on the expected clinically significant error rate across runs. Error classification was performed by a single reviewer without independent second review, a limitation acknowledged in the main text.

\noindent To assess the reproducibility of the primary error classification, a blinded re-annotation sub-study was conducted on a proportional stratified random sample of the 115 divergent patient--question pairs. The pairs were collapsed into three severity strata before sampling, comprising cases classified as annotation error or acceptable and ambiguous (categories A and B), cases classified as partially correct or clinically insignificant error (categories C and D), and cases classified as clinically significant error or pipeline failure (categories E and F). A $40\,\%$ proportional stratified random sample was drawn from each stratum (seed $= 42$), with any rounding remainder assigned to the E+F stratum, yielding $n=46$ cases. Cases were presented in order of patient and question identifier to an independent rater (JS) blinded to the original classification, who applied the same six-category taxonomy and clinical significance threshold used in the primary error classification. Agreement is quantified using Cohen's $\kappa$ for both the collapsed three-stratum scheme and the full six-category scheme, with $95\,\%$ confidence intervals computed by bootstrap ($N_\text{boot} = 10{,}000$). 

\subsection{Citation sufficiency assessment}

\noindent Citation sufficiency was assessed on the same evaluation run used for error classification (overall concordance 77\D8\%). A stratified sample of 96 agentic system responses was drawn, comprising 16 responses from each of six cells defined by the cross-classification of complexity level (Level~1, Level~2, Level~3) and concordance status (concordant, discordant), with the constraint that each of the 48 question templates contributed at least one response. Within each cell, responses were drawn randomly from the available pool after satisfying the template coverage constraint.

\noindent Assessment was performed by two reviewers (KB, CN), who divided the 96 cases between them. For each sampled response, the reviewer was presented with the question text, the adjudicated reference annotation, and the full agent output with inline citation identifiers. Each cited document could be expanded to display the full report text with the cited passage highlighted, documents retrieved but not cited were displayed separately. The reviewer additionally had access to the institutional SAP interface.

\noindent For each response, the reviewer assessed whether the cited sources fully supported the answer on a three-point scale: \textit{fully supported} (cited evidence clearly and completely justified the answer), \textit{partially supported} (cited evidence covered some aspects but was incomplete or insufficient), or \textit{not supported} (cited evidence did not justify the answer). Support was assessed independently of answer correctness. For responses rated as partially or not supported, the reviewer identified whether a relevant document appeared in the retrieved-but-not-cited set, and classified the dominant failure mechanism as one of three types: \textit{reasoning failure} (the relevant document was retrieved but not used or was misinterpreted during synthesis), \textit{retrieval failure} (the relevant document existed in the record but was not returned by the retrieval tools), or \textit{true knowledge gap} (the information required to answer the question was not present in the patient record).
\newpage
\renewcommand{\thesection}{B}
\setcounter{table}{0}
\renewcommand{\thetable}{S\arabic{table}}

\section{Supplementary Tables}
\subsection{Supplementary Table~1: Baseline characteristics of evaluation cohorts}
\noindent Demographic and disease characteristics of patients in the primary evaluation (TUM, $n = 100$) and external validation (MIMIC-IV, $n = 20$) cohorts. Continuous variables are reported as median (IQR), categorical variables as $n$ (\%).

\begin{table}[H]
\centering
\small
\begin{tabular}{lcc}
\hline
\textbf{Characteristic} & \textbf{TUM ($n = 100$)} & \textbf{MIMIC-IV ($n = 20$)} \\
\hline
Age at diagnosis, years          & 65\D1 (57\D3--74\D0) & 66\D5 (58\D8-77\D5)\\
Sex (male), $n$ (\%)             & 63 (63\%)            & 13 (65\%)\\
Documents per patient            & 50\D5 (31\D0--83\D0) & 29\D0 (14\D8--61\D8) \\
Laboratory values per patient    & 1414 (751--2278)      & 3411 (1989--7411) \\
Follow-up span (record), years  & 5\D5 (2\D7 -- 11\D0)& 3\D3 (1\D2 -- 4\D6)\\
\hline
\end{tabular}
\end{table}

\subsection{Supplementary Table~2: Clinical question bank}
\label{tab:question-bank}
\noindent All 48 question templates used for evaluation, grouped by clinical task category and complexity level. Each template includes the question identifier, complexity level, clinical task category, template text with placeholder variables, expected answer format, and scoring method (binary or list-type F1). [\textsc{date}] denotes the date of the last available report. ND\,=\,not documented.

\begin{landscape}
\begin{longtable}{p{0.6cm}cp{2.2cm}p{6.5cm}p{4.2cm}p{1.8cm}}
\hline
\textbf{ID} & \textbf{L} & \textbf{Category} & \textbf{Template text} & \textbf{Answer format} & \textbf{Scoring} \\
\hline
\endfirsthead
\hline
\textbf{ID} & \textbf{L} & \textbf{Category} & \textbf{Template text} & \textbf{Answer format} & \textbf{Scoring} \\
\hline
\endhead
\hline
\endfoot

\multicolumn{6}{l}{\textit{Level~1 -- Simple}} \\[2pt]
Q01 & 1 & Single choice & Is the patient receiving lenalidomide on [\textsc{date}]? & Yes / No / ND / Unclear & Binary \\
Q02 & 1 & Single choice & Is the patient receiving bortezomib on [\textsc{date}]? & Yes / No / ND / Unclear & Binary \\
Q03 & 1 & Single choice & Is the patient receiving daratumumab on [\textsc{date}]? & Yes / No / ND / Unclear & Binary \\
Q04 & 1 & Single choice & Is there documented evidence that a whole-body MRI has been performed? & Yes / No / Unclear & Binary \\
Q05 & 1 & Single choice & Is there documented evidence that a PET-CT has been performed? & Yes / No / Unclear & Binary \\
Q06 & 1 & Single choice & Is there documented evidence that FISH has been performed? & Yes / No / Unclear & Binary \\
Q07 & 1 & Single choice & Has the patient received at least one CD38-antibody therapy? & Yes / No / Unclear & Binary \\
Q08 & 1 & Single choice & Has the patient received at least one IMiD therapy? & Yes / No / Unclear & Binary \\
Q09 & 1 & Single choice & Has the patient received at least one proteasome inhibitor? & Yes / No / Unclear & Binary \\
Q10 & 1 & Single choice & Has the patient received an autologous SCT? & Yes / No / Unclear & Binary \\
Q11 & 1 & Single choice & Has the patient received an allogeneic SCT? & Yes / No / Unclear & Binary \\
Q12 & 1 & Single choice & Has the patient received CAR-T therapy? & Yes / No / Unclear & Binary \\
Q13 & 1 & Single choice & Has the patient received BiTE therapy? & Yes / No / Unclear & Binary \\
Q14 & 1 & Single choice & Has the patient received dialysis? & Yes / No / Unclear & Binary \\
Q15 & 1 & Single choice & Has the patient been diagnosed with sepsis or septic shock? & Yes / No / Unclear & Binary \\
Q16 & 1 & Single choice & Has the patient been mechanically ventilated (invasive or non-invasive)? & Yes / No / Unclear & Binary \\
Q17 & 1 & Single choice & Is renal failure / AKI documented? & Yes / No / ND / Unclear & Binary \\
Q18 & 1 & Single choice & Is clinically relevant anaemia present on [\textsc{date}] (documented or lab-confirmed)? & Yes / No / ND / Unclear & Binary \\
Q19 & 1 & Single choice & Has the patient received red blood cell transfusions? & Yes / No / Unclear & Binary \\
Q20 & 1 & Single choice & Has the patient been diagnosed with pneumonia? & Yes / No / Unclear & Binary \\[6pt]

\multicolumn{6}{l}{\textit{Level~2 -- Medium}} \\[2pt]
Q21 & 2 & Treatment intervals & Which cycle start dates (C*D1) and doses for bortezomib are documented? & $\leq$12$\times$(date; dose; unit) & List-type F1 \\
Q22 & 2 & Treatment intervals & Which cycle start dates (C*D1) and doses for lenalidomide are documented? & $\leq$12$\times$(date; dose; unit) & List-type F1 \\
Q23 & 2 & Treatment intervals & Which melphalan exposures are documented (type, approximate date, dose per episode)? & $\leq$3$\times$(type; date; dose; unit) & List-type F1 \\
Q24 & 2 & Treatment intervals & Which doxorubicin exposures are documented (form, approximate date, cumulative dose per episode)? & $\leq$3$\times$(form; date; dose; unit) & List-type F1 \\
Q25 & 2 & Treatment intervals & In which documented intervals was carfilzomib administered (start--end or ongoing)? & $\leq$3$\times$(start--end / ongoing) & List-type F1 \\
Q26 & 2 & Treatment intervals & In which documented intervals was pomalidomide administered (start--end or ongoing)? & $\leq$3$\times$(start--end / ongoing) & List-type F1 \\
Q27 & 2 & Treatment intervals & In which documented intervals was meropenem administered (start--end or ongoing)? & $\leq$3$\times$(start--end / ongoing) & List-type F1 \\
Q28 & 2 & Single choice & Was toxicity, dose reduction, or discontinuation documented under carfilzomib? If yes, reason? & Yes / No / Unclear + free text & Binary \\
Q29 & 2 & Single choice & Was toxicity, dose reduction, or discontinuation documented under pomalidomide? If yes, reason? & Yes / No / Unclear + free text & Binary \\
Q30 & 2 & Single choice & Best documented response under isatuximab + Pom-Dex? & CR/VGPR/PR/SD/PD / Never / ND / Unclear & Binary \\
Q31 & 2 & Single choice & Best documented response under KRD (carfilzomib--lenalidomide--dexamethasone)? & CR/VGPR/PR/SD/PD / Never / ND / Unclear & Binary \\
Q32 & 2 & Single choice & Best documented response under the documented first-line therapy? & CR/VGPR/PR/SD/PD / Never / ND / Unclear & Binary \\
Q33 & 2 & Single choice & Best documented response after high-dose melphalan? & CR/VGPR/PR/SD/PD / Never / ND / Unclear & Binary \\
Q34 & 2 & Single choice & Best documented response under BiTE? & CR/VGPR/PR/SD/PD / Never / ND / Unclear & Binary \\
Q35 & 2 & Single choice & Best documented response under CAR-T? & CR/VGPR/PR/SD/PD / Never / ND / Unclear & Binary \\
Q36 & 2 & First occurrence & First documented CT report describing new osteolytic lesions? & Date / ND / Unclear & Binary \\
Q37 & 2 & First occurrence & First documented episode of renal failure or dialysis-requiring AKI? & Date / ND / Unclear & Binary \\
Q38 & 2 & Staging & ISS stage on [\textsc{date}] calculated from albumin and $\beta_2$M (last labs $\leq$90\,days prior)? & Score (I/II/III) + date + source & Binary \\[6pt]

\multicolumn{6}{l}{\textit{Level~3 -- Complex}} \\[2pt]
Q39 & 3 & Staging & ECOG score on [\textsc{date}] (last documented $\leq$180\,days, else inferred from explicit functional description)? & Score (0--4) + date + source & Binary \\
Q40 & 3 & Staging & HCT-CI score on [\textsc{date}] (last documented $\leq$365\,days, else derived from comorbidities)? & Score + date + source & Binary \\
Q41 & 3 & Staging & R-ISS stage on [\textsc{date}] (labs $\leq$90\,days; cytogenetics from last available report)? & Stage (I/II/III) + date + source & Binary \\
Q42 & 3 & Staging & R2-ISS stage on [\textsc{date}] (labs $\leq$90\,days; cytogenetics incl.\ 1q status from last available report)? & Stage (I/II/III/IV) + date + source & Binary \\
Q43 & 3 & Single choice & Is the patient triple-class refractory on [\textsc{date}]? (with documented justification) & Yes / No / Unclear + free text & Binary \\
Q44 & 3 & Single choice & Is the patient quadruple-class refractory on [\textsc{date}]? (with documented justification) & Yes / No / Unclear + free text & Binary \\
Q45 & 3 & Single choice & Is the patient eligible for ASCT on [\textsc{date}] based on age, comorbidities, performance status, and organ function? & Yes / No / Unclear + free text & Binary \\
Q46 & 3 & Single choice & Which therapy achieved the best documented response up to [\textsc{date}] (per IMWG criteria)? & Free text (therapy) + CR/VGPR/PR/SD/PD / ND / Unclear & Binary \\
Q47 & 3 & Single choice & Which risks currently dominate: disease progression or therapy toxicity? & Progression / Toxicity / Both / Unclear + free text & Binary \\
Q48 & 3 & Eligibility & Which BCMA-CAR-T eligibility criteria are met / not met / missing on [\textsc{date}]? Is the patient eligible overall? & Criteria table (met/not met/missing) + Yes / No / Unclear & Binary \\

\hline
\end{longtable}
\end{landscape}

\newpage
\subsection{Supplementary Table~3: Pairwise cluster bootstrap significance tests for concordance differences}
\label{tab:pairwise-bootstrap}

\noindent All pairwise comparisons between system configurations, overall and stratified by complexity level. Differences are reported as System~A minus System~B in percentage points; a negative value indicates that System~B outperformed System~A. Confidence intervals and $p$-values were obtained by cluster bootstrap with $N_{\text{boot}} = 10{,}000$ resamples, resampling whole patients with replacement and retaining all questions per sampled patient to account for within-patient correlation. The N column reports the number of patients contributing to each stratum. Bonferroni-corrected $p$-values were computed by multiplying raw $p$-values by the number of pairwise comparisons within each stratum (6 headline comparisons), capped at 1\D0. Significance codes reflect Bonferroni-corrected $p$-values: $^{***}p < 0\D001$; $^{**}p < 0\D01$; $^{*}p < 0\D05$; ns\,$p \geq 0\D05$.

\begin{table}[H]
\centering
\small
\begin{tabular}{llrrrrrl}
\hline
\textbf{Subset} & \textbf{Comparison (A vs B)} & \textbf{N patients} & \textbf{Diff A$-$B (pp)} & \textbf{95\% CI} & \textbf{Raw\,$p$} & \textbf{Bonf.\,$p$} & \\
\hline
\textbf{Overall} \\
Overall & Baseline vs Simple RAG        & 100 & $-$70\D22 & [$-$74\D00, $-$66\D44] & $<$0\D001 & $<$0\D001 & $^{***}$ \\
Overall & Baseline vs Advanced RAG      & 100 & $-$74\D81 & [$-$77\D93, $-$71\D52] & $<$0\D001 & $<$0\D001 & $^{***}$ \\
Overall & Baseline vs Iterative RAG     & 100 & $-$74\D14 & [$-$77\D21, $-$70\D96] & $<$0\D001 & $<$0\D001 & $^{***}$ \\
Overall & Baseline vs Full Context      & 100 & $-$74\D55 & [$-$77\D81, $-$71\D17] & $<$0\D001 & $<$0\D001 & $^{***}$ \\
Overall & Baseline vs Agentic System    & 100 & $-$78\D31 & [$-$81\D37, $-$75\D20] & $<$0\D001 & $<$0\D001 & $^{***}$ \\
Overall & Simple RAG vs Advanced RAG    & 100 & $-$4\D58  & [$-$6\D99, $-$2\D22]  & $<$0\D001 & 0\D0006   & $^{***}$ \\
Overall & Simple RAG vs Iterative RAG   & 100 & $-$3\D92  & [$-$6\D46, $-$1\D38]  & 0\D0024   & 0\D0144   & $^{**}$ \\
Overall & Simple RAG vs Full Context    & 100 & $-$4\D33  & [$-$6\D80, $-$1\D95]  & 0\D0003   & 0\D0018   & $^{***}$ \\
Overall & Simple RAG vs Agentic System  & 100 & $-$8\D09  & [$-$11\D25, $-$4\D93] & $<$0\D001 & $<$0\D001 & $^{***}$ \\
Overall & Advanced RAG vs Iterative RAG & 100 & $+$0\D67  & [$-$0\D64, $+$1\D94]  & 0\D3083   & 1\D000    & ns \\
Overall & Advanced RAG vs Full Context  & 100 & $+$0\D26  & [$-$1\D72, $+$2\D18]  & 0\D7907   & 1\D000    & ns \\
Overall & Advanced RAG vs Agentic       & 100 & $-$3\D51  & [$-$5\D80, $-$1\D26]  & 0\D0025   & 0\D0150   & $^{**}$ \\
Overall & Iterative RAG vs Full Context & 100 & $-$0\D41  & [$-$2\D50, $+$1\D64]  & 0\D6938   & 1\D000    & ns \\
Overall & Iterative RAG vs Agentic      & 100 & $-$4\D17  & [$-$6\D68, $-$1\D78]  & 0\D0012   & 0\D0072   & $^{**}$ \\
Overall & Full Context vs Agentic       & 100 & $-$3\D76  & [$-$6\D13, $-$1\D47]  & 0\D0010   & 0\D0060   & $^{**}$ \\
\hline
Level~1 & Baseline vs Simple RAG        & 100 & $-$78\D90 & [$-$83\D55, $-$74\D05] & $<$0\D001 & $<$0\D001 & $^{***}$ \\
Level~1 & Baseline vs Advanced RAG      & 100 & $-$84\D50 & [$-$88\D65, $-$80\D00] & $<$0\D001 & $<$0\D001 & $^{***}$ \\
Level~1 & Baseline vs Iterative RAG     & 100 & $-$82\D25 & [$-$86\D65, $-$77\D45] & $<$0\D001 & $<$0\D001 & $^{***}$ \\
Level~1 & Baseline vs Full Context      & 100 & $-$84\D90 & [$-$89\D00, $-$80\D60] & $<$0\D001 & $<$0\D001 & $^{***}$ \\
Level~1 & Baseline vs Agentic System    & 100 & $-$86\D00 & [$-$90\D15, $-$81\D60] & $<$0\D001 & $<$0\D001 & $^{***}$ \\
Level~1 & Simple RAG vs Advanced RAG    & 100 & $-$5\D60  & [$-$9\D15, $-$2\D10]  & 0\D0025   & 0\D0150   & $^{**}$ \\
Level~1 & Simple RAG vs Iterative RAG   & 100 & $-$3\D35  & [$-$7\D30, $+$0\D45]  & 0\D0911   & 0\D5466   & ns \\
Level~1 & Simple RAG vs Full Context    & 100 & $-$6\D00  & [$-$9\D60, $-$2\D75]  & 0\D0012   & 0\D0072   & $^{**}$ \\
Level~1 & Simple RAG vs Agentic         & 100 & $-$7\D10  & [$-$11\D65, $-$2\D50] & 0\D0020   & 0\D0120   & $^{**}$ \\
Level~1 & Advanced RAG vs Iterative RAG & 100 & $+$2\D25  & [$-$0\D10, $+$4\D60]  & 0\D0615   & 0\D3690   & ns \\
Level~1 & Advanced RAG vs Full Context  & 100 & $-$0\D40  & [$-$3\D35, $+$2\D50]  & 0\D7878   & 1\D000    & ns \\
Level~1 & Advanced RAG vs Agentic       & 100 & $-$1\D50  & [$-$4\D95, $+$1\D90]  & 0\D3907   & 1\D000    & ns \\
Level~1 & Iterative RAG vs Full Context & 100 & $-$2\D65  & [$-$5\D95, $+$0\D60]  & 0\D1134   & 0\D6804   & ns \\
Level~1 & Iterative RAG vs Agentic      & 100 & $-$3\D75  & [$-$7\D80, $+$0\D40]  & 0\D0733   & 0\D4398   & ns \\
Level~1 & Full Context vs Agentic       & 100 & $-$1\D10  & [$-$4\D35, $+$2\D25]  & 0\D5078   & 1\D000    & ns \\
\hline
\end{tabular}
\end{table}

\begin{table}[H]
\centering
\small
\noindent\textbf{Supplementary Table~3 (continued):} Pairwise cluster bootstrap significance tests for concordance differences
\begin{tabular}{llrrrrrl}
\hline
\textbf{Subset} & \textbf{Comparison (A vs B)} & \textbf{N patients} & \textbf{Diff A$-$B (pp)} & \textbf{95\% CI} & \textbf{Raw\,$p$} & \textbf{Bonf.\,$p$} & \\
\hline
Level~2 & Baseline vs Simple RAG        & 100 & $-$72\D65 & [$-$78\D74, $-$66\D40] & $<$0\D001 & $<$0\D001 & $^{***}$ \\
Level~2 & Baseline vs Advanced RAG      & 100 & $-$75\D89 & [$-$81\D22, $-$70\D41] & $<$0\D001 & $<$0\D001 & $^{***}$ \\
Level~2 & Baseline vs Iterative RAG     & 100 & $-$75\D87 & [$-$80\D90, $-$70\D57] & $<$0\D001 & $<$0\D001 & $^{***}$ \\
Level~2 & Baseline vs Full Context      & 100 & $-$73\D88 & [$-$79\D84, $-$67\D74] & $<$0\D001 & $<$0\D001 & $^{***}$ \\
Level~2 & Baseline vs Agentic           & 100 & $-$77\D76 & [$-$82\D79, $-$72\D40] & $<$0\D001 & $<$0\D001 & $^{***}$ \\
Level~2 & Simple RAG vs Advanced RAG    & 100 & $-$3\D24  & [$-$7\D28, $+$0\D47]  & 0\D0981   & 0\D5886   & ns \\
Level~2 & Simple RAG vs Iterative RAG   & 100 & $-$3\D22  & [$-$7\D11, $+$0\D55]  & 0\D1019   & 0\D6114   & ns \\
Level~2 & Simple RAG vs Full Context    & 100 & $-$1\D23  & [$-$5\D15, $+$2\D63]  & 0\D5401   & 1\D000    & ns \\
Level~2 & Simple RAG vs Agentic         & 100 & $-$5\D11  & [$-$8\D76, $-$1\D56]  & 0\D0058   & 0\D0348   & $^{**}$ \\
Level~2 & Advanced RAG vs Iterative RAG & 100 & $+$0\D01  & [$-$1\D37, $+$1\D36]  & 0\D9859   & 1\D000    & ns \\
Level~2 & Advanced RAG vs Full Context  & 100 & $+$2\D01  & [$-$1\D32, $+$5\D58]  & 0\D2474   & 1\D000    & ns \\
Level~2 & Advanced RAG vs Agentic       & 100 & $-$1\D87  & [$-$4\D57, $+$0\D81]  & 0\D1645   & 0\D9870   & ns \\
Level~2 & Iterative RAG vs Full Context & 100 & $+$1\D99  & [$-$1\D27, $+$5\D34]  & 0\D2354   & 1\D000    & ns \\
Level~2 & Iterative RAG vs Agentic      & 100 & $-$1\D89  & [$-$4\D52, $+$0\D69]  & 0\D1515   & 0\D9090   & ns \\
Level~2 & Full Context vs Agentic       & 100 & $-$3\D88  & [$-$7\D78, $-$0\D32]  & 0\D0418   & 0\D2508   & ns \\
\hline
\textbf{Level 3} \\
Level~3 & Baseline vs Simple RAG        & 90 & $-$46\D11 & [$-$54\D78, $-$37\D67] & $<$0\D001 & $<$0\D001 & $^{***}$ \\
Level~3 & Baseline vs Advanced RAG      & 90 & $-$51\D11 & [$-$59\D11, $-$43\D11] & $<$0\D001 & $<$0\D001 & $^{***}$ \\
Level~3 & Baseline vs Iterative RAG     & 90 & $-$52\D67 & [$-$60\D78, $-$44\D44] & $<$0\D001 & $<$0\D001 & $^{***}$ \\
Level~3 & Baseline vs Full Context      & 90 & $-$52\D89 & [$-$61\D67, $-$44\D22] & $<$0\D001 & $<$0\D001 & $^{***}$ \\
Level~3 & Baseline vs Agentic           & 90 & $-$62\D33 & [$-$71\D00, $-$53\D11] & $<$0\D001 & $<$0\D001 & $^{***}$ \\
Level~3 & Simple RAG vs Advanced RAG    & 90 & $-$5\D00  & [$-$9\D56, $-$0\D55]  & 0\D0295   & 0\D1770   & ns \\
Level~3 & Simple RAG vs Iterative RAG   & 90 & $-$6\D56  & [$-$11\D44, $-$1\D89] & 0\D0078   & 0\D0468   & $^{**}$ \\
Level~3 & Simple RAG vs Full Context    & 90 & $-$6\D78  & [$-$13\D22, $-$0\D22] & 0\D0420   & 0\D2520   & ns \\
Level~3 & Simple RAG vs Agentic         & 90 & $-$16\D22 & [$-$25\D11, $-$7\D11] & 0\D0004   & 0\D0024   & $^{***}$ \\
Level~3 & Advanced RAG vs Iterative RAG & 90 & $-$1\D56  & [$-$4\D56, $+$1\D22]  & 0\D2847   & 1\D000    & ns \\
Level~3 & Advanced RAG vs Full Context  & 90 & $-$1\D78  & [$-$6\D44, $+$3\D00]  & 0\D4521   & 1\D000    & ns \\
Level~3 & Advanced RAG vs Agentic       & 90 & $-$11\D22 & [$-$18\D44, $-$4\D00] & 0\D0020   & 0\D0120   & $^{**}$ \\
Level~3 & Iterative RAG vs Full Context & 90 & $-$0\D22  & [$-$5\D33, $+$5\D11]  & 0\D9310   & 1\D000    & ns \\
Level~3 & Iterative RAG vs Agentic      & 90 & $-$9\D67  & [$-$16\D67, $-$2\D67] & 0\D0082   & 0\D0492   & $^{**}$ \\
Level~3 & Full Context vs Agentic       & 90 & $-$9\D44  & [$-$16\D22, $-$2\D67] & 0\D0054   & 0\D0324   & $^{**}$ \\
\hline
\end{tabular}
\end{table}

\newpage
 \subsection{Supplementary Table~4: Run-to-run stability across system configurations}
  \label{tab:stability}
  \noindent Concordance estimates across ten independent evaluation runs for each system configuration on the primary evaluation cohort (TUM, $n = 469$). Results are reported as mean $\pm$ standard
  deviation across runs, overall and stratified by complexity level. Individual run results are listed to characterise the distribution.

\begin{table}[H]
\centering
\small
\begin{tabular}{lcccc}
\hline
\textbf{System} & \textbf{Overall} & \textbf{Level 1} & \textbf{Level 2} & \textbf{Level 3} \\
\hline
Baseline             & 1\D3\, $\pm$\, 0\D5\%  & 0\D1\, $\pm$\, 0\D2\%  & 1\D8\, $\pm$\, 1\D0\%  & 2\D8\, $\pm$\, 1\D1\%  \\
Simple RAG System    & 71\D5\, $\pm$\, 0\D9\% & 79\D0\, $\pm$\, 1\D7\% & 74\D4\, $\pm$\, 0\D9\% & 48\D9\, $\pm$\, 2\D7\% \\
Advanced RAG System  & 76\D1\, $\pm$\, 1\D7\% & 84\D7\, $\pm$\, 1\D5\% & 77\D7\, $\pm$\, 1\D3\% & 53\D9\, $\pm$\, 4\D5\% \\
Iterative RAG System & 75\D4\, $\pm$\, 2\D2\% & 82\D4\, $\pm$\, 2\D0\% & 77\D6\, $\pm$\, 3\D0\% & 55\D4\, $\pm$\, 3\D0\% \\
Full Context         & 75\D8\, $\pm$\, 0\D5\% & 85\D0\, $\pm$\, 1\D0\% & 75\D7\, $\pm$\, 1\D1\% & 55\D7\, $\pm$\, 1\D8\% \\
Agentic System       & 79\D6\, $\pm$\, 1\D1\% & 86\D2\, $\pm$\, 1\D1\% & 79\D5\, $\pm$\, 2\D4\% & 65\D1\, $\pm$\, 2\D2\% \\
\hline
\end{tabular}
\end{table}

\subsection{Supplementary Table~5: System ablations}
\label{tab:ablations}
\noindent Concordance across ablated system configurations on the primary evaluation cohort (TUM, $n = 469$). Values are mean concordance (\%) across the number of independent runs indicated, overall and stratified by complexity level. Each row removes or replaces one or more components of the full agentic system. The unfiltered skill library condition loads all 41 skill modules unconditionally rather than selecting by query content. Configurations without a skill library use the same fixed baseline prompt as the comparator systems, without the agent-specific reasoning skills.

\begin{table}[H]
\centering
\small
\begin{tabular}{lrcccc}
\hline
\textbf{Configuration} & \textbf{Overall} & \textbf{Level 1} & \textbf{Level 2} & \textbf{Level 3} \\
\hline
Standard configuration & 79\D6\% & 86\D2\% & 79\D5\%&65\D1\% \\\hline

No deterministic clinical scoring tools &79\D6\%&85\D1\%&80\D0\%&66\D7\%\\

No type and date filters in retrieval tools & 79\D5\% & 87\D7\% & 78\D1\%&63\D8\%\\
Full skill library, unfiltered               &  79\D3\% & 87\D7\% & 78\D7\% & 62\D0\% \\
No structured memory state        & 79\D2\% & 85\D9\% & 79\D0\% & 64\D9\% \\
Reactive tool selection (no pre-planned use) &  79\D2\% & 86\D6\% & 78\D7\% & 63\D8\% \\

No skill library                             &  76\D6\% & 86\D0\% & 77\D9\% & 53\D3\% \\
\hline
\end{tabular}
\end{table}

\subsection{Supplementary Table~6: System performance across language model backbones}
\label{tab:backbone-sensitivity}
\noindent All four system configurations were evaluated under each backbone model on the primary evaluation cohort (469 patient-question pairs, TUM). Each model was deployed locally via vLLM using the inference settings recommended by the respective developer. Weight quantization reflects the precision of the loaded model checkpoint (MXFP4 or FP8), Qwen3-Next-80B-A3B-Instruct additionally used an FP8 KV cache. Overall concordance (\%) is reported as mean across ten runs. The model used in the primary analysis is indicated ($^{*}$).

\begin{table}[H]
\centering
\small
\begin{tabular}{lrcccccc}
\hline
\textbf{Model} & \textbf{Params} & \textbf{Quantization}& \textbf{Simple RAG} & \textbf{Iterative RAG} & \textbf{Full Context} & \textbf{Agentic}  \\
\hline
gpt-oss-120b$^{*}$ & 120B & MXFP4 & 71\D5\% & 75\D4\% & 75\D8\% & 79\D6\%  \\
GLM-4.5-Air & 110B & FP8 & 64\D9\% & 68\D2\% & 72\D0\% & 77\D9\%  \\
Qwen3-Next-80B-A3B-Instruct  & 81B & FP8& 75\D0\% & 76\D5\% & 76\D1\% & 76\D8\% \\
gemma-4-31B-it  & 33B & FP8 & 74\D7\% & 82\D4\% & 81\D3\% & 81\D6\%\\
gpt-oss-20b & 21B & MXFP4 & 67\D9\% & 72\D1\% & 68\D4\% & 77\D4\% \\

\hline
\end{tabular}
\end{table}

\subsection{Supplementary Table~7: Concordance by clinical task category and system configuration}
\label{tab:category-breakdown}
\noindent Concordance stratified by clinical task category (single choice, treatment intervals, first occurrence, staging, eligibility) for each system configuration on the primary evaluation cohort
(TUM, $n = 469$). Values are mean concordance across ten independent runs with 95\% bootstrap confidence intervals (cluster bootstrap, 10{,}000 resamples, patient-level resampling). Category sample sizes: single choice $n =
326$, treatment intervals $n = 70$, first occurrence $n = 20$, staging $n = 45$, eligibility $n = 8$. Results for the eligibility category are based on a small sample and should be considered hypothesis-generating only.

\begin{table}[H]
\centering
\small
\begin{tabular}{lccccc}
\hline
\textbf{System} & \textbf{Single choice} & \textbf{Treatment intervals} & \textbf{First occurrence} & \textbf{Staging} & \textbf{Eligibility} \\
\hline
Baseline        & 0\D1 [0\D0--0\D2] & 2\D6 [1\D7--3\D6] & 0\D0 [0\D0--0\D0] & 8\D4 [6\D2--10\D9] & 0\D0 [0\D0--0\D0] \\
Simple RAG      & 76\D9 [76\D3--77\D5] & 74\D2 [73\D0--75\D2] & 38\D0 [35\D5--40\D5] & 55\D6 [54\D7--56\D4] & 1\D2 [0\D0--3\D8] \\
Advanced RAG    & 81\D8 [81\D0--82\D8] & 74\D6 [73\D5--75\D6] & 52\D5 [47\D5--57\D0] & 56\D4 [55\D3--57\D6] & 25\D0 [16\D2--33\D8] \\
Iterative RAG   & 80\D9 [79\D1--82\D1] & 74\D6 [73\D0--76\D0] & 52\D0 [48\D5--55\D5] & 55\D3 [53\D3--57\D1] & 32\D5 [26\D2--40\D0] \\
Full Context    & 83\D5 [82\D9--84\D0] & 70\D9 [70\D1--71\D7] & 40\D0 [37\D5--42\D5] & 53\D6 [52\D0--55\D3] & 21\D2 [13\D8--28\D7] \\
Agentic System  & 86\D4 [85\D7--87\D0] & 76\D1 [74\D9--77\D3] & 50\D5 [46\D0--55\D0] & 52\D4 [50\D4--54\D7] & 58\D8 [53\D8--63\D7] \\
\hline
\end{tabular}
\end{table}

\subsection{Supplementary Table~8: Execution characteristics by complexity level and system configuration}
\label{tab:execution-profile}
\noindent Wall-clock time per question on the primary evaluation cohort (TUM, $n$ = 469 patient-question pairs from 100 patients), stratified by complexity level and system configuration. Unless otherwise indicated, all values are median (IQR) across all patient-question pairs from a single-concurrency timing run on a single NVIDIA H200 GPU. Skills selected, tool calls, and document retrieval metrics are reported for the agentic system only.
\begin{table}[H]
\centering
\small
\begin{tabular}{lcccc}
\hline
 & \textbf{Level~1} & \textbf{Level~2} & \textbf{Level~3} & \textbf{Overall} \\
\textbf{Metric} & \textit{($n=200$)} & \textit{($n=179$)} & \textit{($n=90$)} & \textit{($n=469$)} \\
\hline
\multicolumn{5}{l}{\textit{Agentic system}} \\
Skills selected per question             & 6\D0 (6\D0--7\D0)   & 6\D0 (6\D0--7\D0)   & 8\D0 (7\D0--9\D0)   & 7\D0 (6\D0--7\D0)   \\
Tool calls per question                  & 2\D0 (1\D0--3\D0)   & 3\D0 (2\D0--4\D0)   & 4\D0 (3\D0--4\D0)   & 3\D0 (2\D0--4\D0)   \\
Documents retrieved                & 15\D0 (7\D5--26\D0) & 27\D0 (14\D0--49\D0)& 27\D0 (16\D0--46\D0)& 20\D0 (10\D0--39\D0)\\
Documents cited                    & 2\D0 (2\D0--2\D0)   & 2\D0 (2\D0--2\D0)   & 2\D0 (2\D0--4\D0)   & 2\D0 (2\D0--2\D0)   \\
\quad of which $\geq$3 documents (\%)       & 5\D9                 & 9\D5                 & 47\D4                & 15\D2                \\
Wall-clock time per question, s          & 18\D5 (16\D7--20\D6)& 21\D3 (19\D4--23\D8)& 28\D7 (26\D3--34\D2)& 20\D9 (18\D2--25\D2)\\
\multicolumn{5}{l}{\textit{Iterative RAG}} \\
Wall-clock time per question, s          & 6\D0 (5\D0--7\D3)   & 6\D4 (5\D1--8\D6)   & 12\D3 (7\D2--23\D4) & 6\D6 (5\D3--9\D5)   \\
\multicolumn{5}{l}{\textit{Full Context}} \\
Wall-clock time per question, s          & 4\D7 (3\D8--5\D6)   & 5\D0 (4\D0--6\D0)   & 6\D0 (5\D1--6\D9)   & 5\D0 (4\D0--6\D0)   \\
\multicolumn{5}{l}{\textit{Simple RAG}} \\
Wall-clock time per question, s          & 2\D2 (1\D8--2\D8)   & 2\D2 (1\D8--2\D8)   & 2\D7 (2\D2--3\D1)   & 2\D3 (1\D8--2\D9)   \\
\hline
\end{tabular}
\end{table}

\subsection{Supplementary Table~9: Patient stratification by record length}
\label{tab:context-length-bins}
\noindent Patients were stratified by total clinical record length (character count of the full concatenated record) into four bins for the context-length sensitivity analysis. Three bins span the lower 90th percentile of the distribution (Q1--Q3), defined at the 33rd and 67th percentiles, the fourth bin (Q4) comprises the top decile. Evaluable pairs are summed across all complexity levels.

\begin{table}[H]
\centering
\small
\begin{tabular}{lcccc}
\hline
\textbf{Bin} & \textbf{Character range} &  \textbf{Patients ($n$)} & \textbf{Evaluable pairs ($n$)} \\
\hline
Q1 ($\leq$p33)              & $\leq$127k          & 33 & 155 \\
Q2 (p33--p67)    & 127k--282k                    & 34 & 160 \\
Q3 (p67--p90)    & 282k--541k                     & 23 & 107 \\
Q4 ($>$p90)      & $>$541k                       & 10 &  47 \\
\hline
Total            & $\leq$1{,}076k                & 100 & 469 \\
\hline
\end{tabular}
\end{table}
\newpage
\renewcommand{\thesection}{C}
\setcounter{figure}{0}
\renewcommand{\thefigure}{S\arabic{figure}}

\section{Supplementary Figures}

\subsection{Supplementary Figure~1: Accuracy by patient sex, vital status, and age group}
\label{fig:run-distributions}

\begin{figure}[!h]
\centering

\includegraphics[width=0.9\columnwidth]{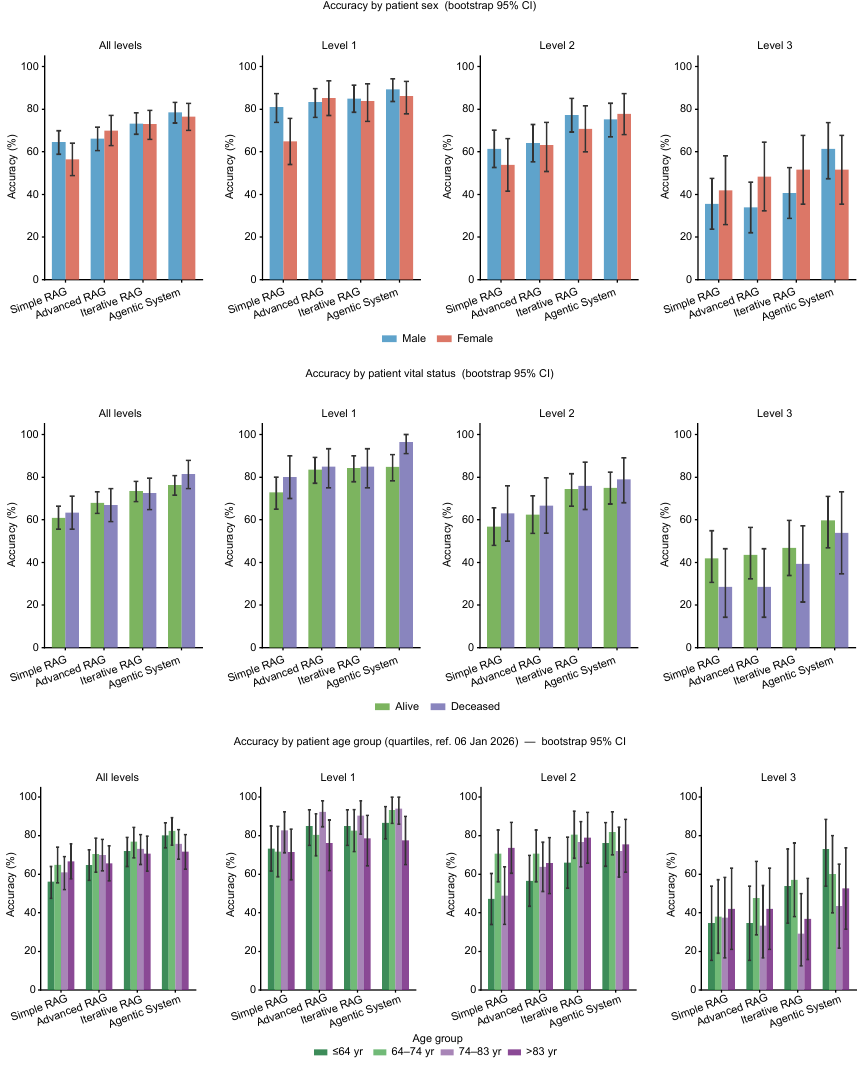}
    \caption{Concordance with expert consensus stratified by patient sex (male, female), vital status (alive, deceased), and age group at the reference date (quartiles: $\leq$64, 64--74, 74--83, $>$83 years), shown overall and by complexity level for all four system configurations. Bootstrap 95\% confidence intervals were computed by cluster bootstrap resampling at the patient level ($N_\text{boot} = 10{,}000$). }
\label{fig:supp-sex-vital-age}
\end{figure}

\newpage
\subsection{Supplementary Figure~2: Accuracy by question template across different systems}
\label{fig:question-wise}

\begin{figure}[!h]
\centering

\includegraphics[width=\columnwidth]{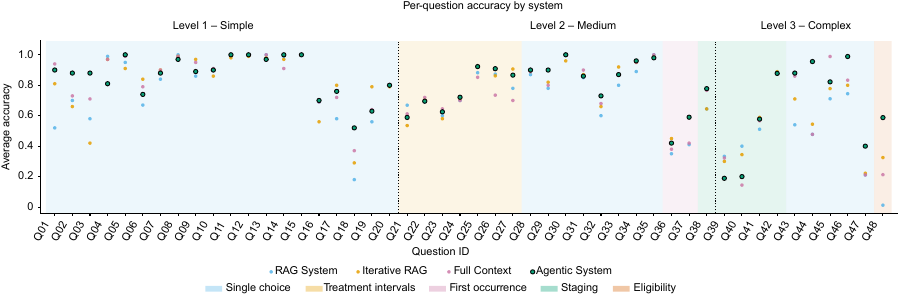}
    \caption{Per-question concordance averaged across ten independent evaluation runs for each of the 48 clinical question templates, stratified by system configuration and clinical task category (single choice, treatment intervals, first occurrence, staging, eligibility). Templates are ordered by complexity level (Level~1, Level~2, Level~3) and grouped by task category within each level. }
\label{fig:supp-per-question}
\end{figure}
\end{document}